\documentclass[12pt]{article}

\usepackage{etex}
\usepackage[hmargin=3cm,vmargin=3cm]{geometry}
\usepackage[ampersand]{easylist}
\usepackage{amssymb}
\usepackage{amsbsy}
\usepackage{amsmath,epsfig}
\usepackage{bbm}
\usepackage{bm}
\usepackage{amsfonts}
\usepackage{array}
\usepackage[usenames]{color}
\usepackage{colortbl}
\usepackage{cite}
\usepackage{float}
\usepackage{framed}
\usepackage{graphicx}
\usepackage{graphics}
\usepackage{lineno,hyperref} \modulolinenumbers[5]
\usepackage{lscape}
\usepackage{makecell}
\usepackage{mathtools}
\usepackage{multirow, booktabs}
\usepackage[normalem]{ulem}
\usepackage{pgf}
\usepackage{tikz}
\usepackage{rotating}
\usepackage{sectsty}
\usepackage{soul}
\usepackage{subfigure}
\usepackage{times}
\usepackage[colorinlistoftodos,backgroundcolor=yellow,textsize=scriptsize,linecolor=black]{todonotes}
\usepackage{verbatim}
\usepackage{upgreek}
\usepackage{url}
\usepackage[all]{xy}
\usepackage{hyperref}
\hypersetup{
    bookmarks=true,         
    unicode=false,          
    pdftoolbar=true,        
    pdfmenubar=true,        
    pdffitwindow=false,     
    pdfstartview={FitH},    
    pdftitle={AMOGAPE},         
    pdfauthor={Svendsen, Martino, Camps-Valls}, 
    pdfsubject={AMOGAPE},       
    pdfcreator={Svendsen, Martino, Camps-Valls},  
    pdfproducer={Svendsen, Martino, Camps-Valls}, 
    pdfkeywords={keyword1} {key2} {key3}, 
    pdfnewwindow=true,      
    colorlinks=true,        
    linkcolor=gray,         
    citecolor=gray,         
    filecolor=blue,         
    urlcolor=blue           
}

\newcommand{\cmmnt}[1]{\ignorespaces}

\def\x{{\mathbf x}}
\def\X{{\mathbf X}}
\def\K{{\mathbf K}}

\def\Y{{\mathbf Y}}
\def\z{{\mathbf z}}
\def\f{{\mathbf f}}

\def\y{{\mathbf y}}
\def\X{{\mathbf X}}
\def\Y{{\mathbf Y}}

\definecolor{MYCOLOR0}{rgb}{0.92,0.92,0.92}
\definecolor{MYCOLOR}{rgb}{1,1,0}
\definecolor{MYCOLOR2}{rgb}{0.5,1,0.5}
\definecolor{MYCOLOR3}{rgb}{0.88,1,1}

\usetikzlibrary{arrows,automata,positioning}
\usetikzlibrary{shapes,arrows,positioning}
\tikzset{
    >=stealth',
    punkt/.style={
           rectangle,
           rounded corners,
           draw=black, very thick,
           text width=6.5em,
           minimum height=2em,
           text centered},
    pil/.style={
           ->,
           thick,
           shorten <=2pt,
           shorten >=2pt,}
}

\usepackage{fancyhdr}

\begin{document}

\hypersetup{%
    ,urlcolor=blue
    ,citecolor=blue
    ,linkcolor=blue
    }

\lhead{ \scriptsize \textcopyright ELSEVIER. ACCEPTED FOR PUBLICATION IN PATTERN RECOGNITION. DOI 10.1016/j.patcog.2019.107103 }
\lfoot{\footnotesize $^\ast$Corresponding author, daniel.svendsen@uv.es }
\thispagestyle{fancy}
{\LARGE \flushleft Active Emulation of Computer Codes with Gaussian Processes -- Application to Remote Sensing}\\ \vspace{0.3cm}

{\large \noindent {Daniel Heestermans Svendsen$^{a,\ast}$}, {Luca Martino$^b$}, {Gustau Camps-Valls$^a$ } }\\ \vspace{-0.1cm}

{\footnotesize \noindent \textit{$^a$Image Processing Lab (IPL), Universitat de Val\`encia, C/ Cat. Jos\'e Beltr\'an, 2. 46980 Paterna, Spain.}\\
\noindent  \textit{$^b$Dep. Signal Processing, Universidad Rey Juan Carlos (URJC), Camino del Molino 5, 28943 Fuenlabrada, Spain} }

\begin{abstract}
Many fields of science and engineering rely on running simulations with complex and computationally expensive models to understand the involved processes in the system of interest. 
Nevertheless, the high cost involved hamper reliable and exhaustive simulations. Very often such codes incorporate heuristics that ironically make them less tractable and transparent. 
This paper introduces an {active learning} methodology for adaptively constructing surrogate models, i.e. {\em emulators}, of such costly computer codes in a multi-output setting. 
The proposed technique is sequential and adaptive, and is based on the optimization of a suitable acquisition function. It aims to achieve accurate approximations, model tractability, as well as compact and expressive simulated datasets. 
In order to achieve this, the proposed Active \cmmnt{automatic} Multi-Output Gaussian Process Emulator (AMOGAPE) combines the predictive capacity of Gaussian Processes (GPs) with the design of an acquisition function that favors sampling in low density and fluctuating regions of the approximation functions. 
Comparing different acquisition functions, we illustrate the promising performance of the method for the construction of \cmmnt{accurate and compact} emulators with toy examples, as well as for a widely used remote sensing transfer code.

\vspace{0.5cm}

\end{abstract}




\section{Introduction}\label{sec:1}

In many areas of science and engineering, systems are analyzed by running computer code simulations which act as convenient approximations of reality.
They allow us to simulate many different systems of interest and characterize the involved processes, such as turbulence or energy transfer, and their interactions and relevance. Depending on the body of literature, they are known as physics-based or mechanistic models, or simply {\em simulators}~\cite{Santer03,Wescott13}. 
Two important limitation are associated with simulators. The first, and perhaps the most important problem of these computer codes, is their often high computational cost, which hampers reliable and exhaustive simulations. This limits the representativity of the simulations, which in turn makes numerical or statistical inversion a hard problem. Secondly, since computer codes rely on decades of intensive development and parametrizations, they often include heuristics that improve accuracy but ironically make them less mathematically tractable and transparent. 

\paragraph{Emulation for forward models.} In the last decade, a field collectively known as {\em surrogate modeling} or {\em emulation} has emerged as an efficient alternative: emulators try to mimic costly computer codes with machine learning models. The field of emulation has received attention from subfields of statistical signal processing and machine learning~\cite{gorissen2009sequential,gorissen2010surrogate,Kennedy2001,OHagan2006,oakley1999bayesian}. 
In order to construct an emulator, we need a simulated dataset which is made by evaluating the computer code in different input points. The problem of choosing these points, {for which this paper presents an active learning algorithm} \cite{WANG20112375,YANG2018401}, is treated in different parts of the statistics and machine learning literature. A non-exhaustive overview is given below.


\paragraph{Related work.} The problem at hand is closely related to that of Design of Experiments (DOE), where one seeks a set of input values which best allows one to determine the relationship between inputs and outputs. 
Between the algorithms that will be reviewed in this section, there are key some differences between types of algorithms that it would be beneficial to clarify first:
\begin{itemize}
    \item {\em Sequential vs. non-sequential} refers to whether the algorithm needs to know a priori how many input points to choose. Non-sequential or one-shot algorithms need this information, while sequential algorithms can simply run until some time limit or accuracy criteria is met, which is a favourable property. Between the two approaches lie the batch-sequential algorithms.
    \item {\em Continuous vs. discrete sampling} refers to whether an algorithm aims to either choose input points in a continuous space or choose among a finite set of points. A large part of the literature deals with the latter problem, and relies on greedy and MCMC algorithms to choose points according to some criterion. Many of the methods proposed in the literature can be easily adapted from continuous to discrete sampling and vice versa.
\end{itemize}
Among the most popular criteria are maximum entropy \cite{shewry1987maximum}, maximizing distance to nearest neighbour \cite{johnson1990minimax}, and minimizing integrated root mean squared error \cite{sacks1989designs}. These criteria have been implemented for construction of GP emulators in both a sequential and batch-sequential way \cite{loeppky2010batch}. An interesting approach in this field is the Bayesian Experimental Design (BED) which assumes a probabilistic model of the observed data and defines a so-called utility function based on the posterior of the model parameters. The approach then aims to maximize the mean of the utility. Recent relevant work can be found in \cite{Ryan16,Gillespie18}. A great deal of DOE methods, even the sequential ones, do not assume the ability to \textit{query} a system, due to the way experiments are carried out. 

The field of Active Learning (AL), on the other hand, builds on the premise that we can query a system and thus learn something about it in each iteration~\cite{mitra2004segmentation,tuia2012remote}. Building an emulator sequentially is a problem that fits directly into this category. 
The algorithms in the AL literature concerned with GP regression often employ criteria based on predictive variance \cite{seo2000gaussian} and entropy \cite{wang2018adaptive}. Other algorithms are based on triangulation of the input space \cite{ajdari2014adaptive} or gridding \cite{busby2009hierarchical}, followed by a ranking of each triangle or cell.
Greedily searching for candidate points which have maximum distance to their nearest neighbours \cite{wu2019active} has also proven effective. Furthermore, when the input set is comprised of finite discrete values, interesting criteria like mutual information have been employed with success \cite{krause2008near}.

\begin{figure}[t!]
\centerline{\includegraphics[width=\textwidth]{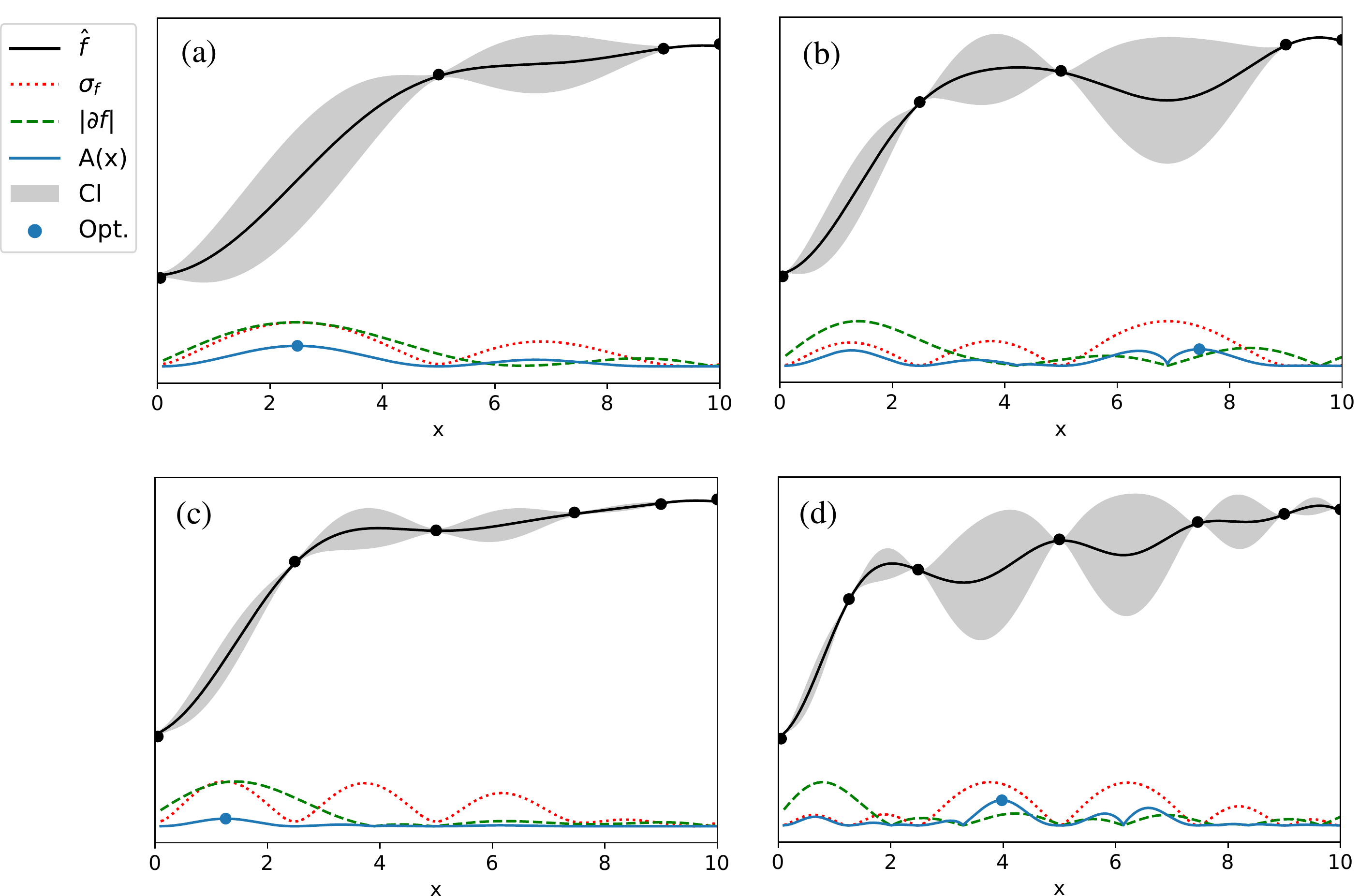}}
\vspace{-0.25cm}
\caption{The presented method optimizes the selection of most informative points to approximate an arbitrary multidimensional function iteratively. The example shows the first four iterations in a 1D case. Starting from 4 points, a GP interpolator is built from which some valuable information is derived (the predictive variance -green- and the gradient -red-) and then combined in an acquisition function (blue) that proposes the next point to sample (blue dot). The acquisition function admits many general forms and trades off geometry and diversity terms to account for attractiveness in the sample space.}
\label{fig:example}
\end{figure}

\paragraph{Our Contribution: Active emulation as a step forward.}

In this paper, we introduce a methodology
for developing efficient machine learning emulators of costly physical models based on active emulation.
An active learning framework is developed that \textit{sequentially} chooses informative input points, learning about about the underlying function as the algorithm progresses. This active emulation methodology is based on the notion of an {\it acquisition function} which can be optimized through gradient-based techniques, mirroring approaches in Bayesian Optimization~\cite{snoek2012practical}. The goal is to construct an accurate emulator with as few runs of the computer code as possible.

Given a set of initial datapoints, the emulator is built through the online addition of new nodes\footnote{ In the following, the words \textit{node} and \textit{datapoint} will be used intercheangebly. }, maximizing the acquisition function at each iteration. The acquisition function is constructed to incorporate (a) geometric information of the costly, analytically intractable function $\f$, and (b) information about the distribution of the current nodes. By using Gaussian processes we can derive both terms analytically, and for multiple outputs at once. The reasoning is that areas of high variability in $\f(\x)$ requires the addition of more information, as has also been noted in \cite{busby2009hierarchical}. In \cite{marmin2018warped} the predictive variance of the gradient norm of a GP is used as a sampling criteria, which is a less straightforward approach than just using the gradient directly as done here.
Similarly, regions with a small concentration of nodes requires the introduction of new nodes in order to fill the space (simple exploration, space filling without taking into account the geometrical features of $\f(\x)$). We show how to define such an acquisition function in a multi-output setting.
Figure~\ref{fig:example} shows an illustrative example of the building blocks of the active emulation methodology presented here.

The developed methodology of constructing emulators is sequential and searches a continuous input-space, leading to emulators that are {\em accurate}, so they can be taken as a faithful representation of the physical models and codes, {\em compact}, and {\em parsimonious}, as a minimal number of informative points is selected, and {\em general-purpose} since it is based on properties of Gaussian processes like uncertainty and gradients that can be obtained for any differentiable covariance function. 
{This paper builds on of the preliminary work in \cite{svendsen2018multioutput}, extending it in several directions. 
A general framework is provided before describing some specific implementations, extending the study proposing the use of a range of different acquisition functions. A theoretical demonstration of the utility of a gradient term in active sampling and emulation is also given  (see~\ref{appendix}, for instance). Finally, more thorough experimental results are provided, with more examples and challenging model comparisons, and a more advanced use case.}

\paragraph{Structure of the paper.}
The remainder of the paper is organized as follows. We first define active \cmmnt{automatic} emulation and establish the notation in Section \ref{sec:2}. Then, the GP-based active \cmmnt{automatic} emulation framework is presented in Section \ref{sec:3}. The framework defines a general-purpose acquisition function built on optimal search of diversity and uncertainty criteria. Experimental results in synthetic and challenging real problems illustrate the capabilities in Section \ref{sec:4}. We will pay special attention to the field of remote sensing, where computer codes, called radiative transfer models (RTMs), are widely used and pose challenges to the design of accurate and compact emulators. Having access to an exhaustive ground truth allows us to analyze performance in terms of convergence and accuracy. We conclude in Section \ref{sec:5} with some remarks and an outline of future work.

\section{Active Emulation} \label{sec:2}

In this section we describe the generic active emulation (AE) method for a complex system denoted as ${\bf f}(\x)$, e.g., an expensive RTM model.
We first fix the notation, then present the processing scheme. 
Consider a $D$-dimensional bounded input space $\mathcal{X}$, i.e., $\x \in\mathcal{X}\subset \mathbb{R}^D$. Furthermore, let  ${\bf f}(\x) \colon \mathcal{X}\to \mathbb{R}^P$ denote a complex system with $P$ outputs. 
Finally, $t\in \mathbb{N}$ denotes the index of the AE algorithm, and $m_t$ the number of datapoints $\{\x_k,\y_k\}_{k=1}^{m_t}$ used by the algorithm at iteration $t$, where 
\begin{eqnarray}
\label{datapoints}
{\bf y}_k=\f(\x_k),
\end{eqnarray} 
 where ${\bf y}_k=[y_{1,k},\ldots,y_{P,k}]^{\intercal}$ and $k=1,\ldots,m_t$. Thus, given an input matrix of nodes, $\X_t=[\x_1,\cdots,\x_{m_t}]$ of dimension $D\times m_t$, we have a $P\times m_t$ matrix of outputs,  ${\bf Y}_t=[{\bf y}_1,\ldots, {\bf y}_{m_t}]$. At each iteration $t$, given the datapoints $\{\x_k,\y_k\}_{k=1}^{m_t}$, the AE method constructs an interpolating function $\widehat{\f}_t(\x)$. 
 Then, an acquisition function $A_{t}(\x)\colon \mathbb{R}^D \rightarrow  \mathbb{R}$ is built in order to suggest which regions of the space require additional nodes. That is, an optimization step is performed for obtaining the next input $\x_{m_t+1}$: 
\begin{equation}
\x_{m_t+1}=\arg\max_{\x\in \mathcal{X}} A_{t}(\x).
\end{equation}
The dataset is updated accordingly, $\X_{t+1} =[\X_t, \x_{m_t+1}]$, ${\Y}_{t+1}=[{\bf Y}_t, {\bf y}_{m_t+1}=\f(\x_{m_t+1})]$  adding a new node, and we set $m_{t+1}=m_t+1$ and $t\leftarrow t +1$. The procedure is repeated until a stopping condition is met. One possibility is to stop the algorithm when a pre-established maximum number of points $M$ (determined by the available computational resources) has been included. Theoretically, the user could stop the algorithm when a least a precision error $\epsilon>0$ is achieved, $\|\f(\x)-\widehat{\f}_{{ t}}(\x)\|\leq \epsilon$.
{However, since $\f(\x)$ is costly and analytically intractable in general\footnote{The system $\y=\f(\x)$ is a black-box mapping, linking the inputs $\x$ with the outputs $\y$. At each new input $\x'$, the system returns $\y'=\f(\x')$,  but does so by way of a computer code which is too complex and slow to lend itself to exhaustive analysis across the input space.}, one cannot evaluate and/or approximate the associated error $\|\f(\x)-\widehat{\f}_{{t}}(\x)\|$.} A practical alternative is to stop the AE method when $\|\widehat{\f}_t(\x)-\widehat{\f}_{t-1}(\x)\|\leq \epsilon'$ for some $\epsilon'>0$. 
Figure~\ref{fig:example} shows a graphical representation of a generic AE procedure. 
Table \ref{tab:notation} summarizes the main notation of the work. Table \ref{alg:GenAE} shows in details the steps of a generic AE algorithm.
\newline
Note that the goal is either to sequentially  construct an emulator able to obtain a pre-established error in approximation with the smallest number of nodes possible or, more commonly, the best possible emulator built with a pre-established maximum number of nodes (given some starting points).  
The constructed emulator will be used for further applications for the interested users, researchers and practitioners. We do not consider time or computational restrictions in the construction stage. Furthermore,  our approach is particularly useful when the underlying function is very costly, i.e. when the cost of evaluating this function is significantly greater than the application of one iteration of the proposed algorithm.

\begin{table}[!t]
\centering
\caption{\normalsize Main notation of the work.}
\vspace{0.1cm}
	\begin{tabular}{|l|l||c|l|}
	 \hline
  \cellcolor{MYCOLOR0} $t\in \mathbb{N}$ &  \multicolumn{3}{l|}{Iteration index of the active \cmmnt{automatic} emulator.} \\
  \cellcolor{MYCOLOR0} $m_t$ &  \multicolumn{3}{l|}{Number of data points at the $t$-th iteration.} \\
    \cellcolor{MYCOLOR0} $\{\x_k,\y_k\}_{k=1}^{m_t}$ &  \multicolumn{3}{l|}{Data points at the $t$-th iteration.} \\
      \cellcolor{MYCOLOR0} ${\bf x}=[x_1,\ldots,x_D]^{\intercal}\in\mathcal{X}\subset \mathbb{R}^D$ & \multicolumn{3}{l|}{Input variable.} \\
  \cellcolor{MYCOLOR0} ${\bf y}=[y_1,\ldots,y_P]^{\intercal}$ & \multicolumn{3}{l|}{Outputs.} \\
   \cellcolor{MYCOLOR0} $\X_t=[\x_1,\cdots,\x_{m_t}]$ &  \multicolumn{3}{l|}{$D\times m_t$ input matrix.} \\
   \cellcolor{MYCOLOR0} ${\bf Y}_t=[{\bf y}_1,\ldots, {\bf y}_{m_t}]$ &  \multicolumn{3}{l|}{$P\times m_t$ output matrix.} \\
     \cellcolor{MYCOLOR0} ${\bf y}=f(\x)\colon \mathcal{X}\to \mathbb{R}^P$ & \multicolumn{3}{l|}{Unknown function/forward model linking $\x$ with $\y$.} \\
  \cellcolor{MYCOLOR0} ${\bf \widehat{y}}=\widehat{\f}_t(\x) \colon \mathcal{X}\to \mathbb{R}^P$ &  \multicolumn{3}{l|}{Interpolator the $t$-th iteration using $\{\x_k,\y_k\}_{k=1}^{m_t}$.} \\
   \cellcolor{MYCOLOR0} $A_{t}(\x) \colon \mathcal{X}\to\mathbb{R}$ &  \multicolumn{3}{l|}{Acquisition function at the $t$-th iteration.} \\ 
     \hline
        \hline
     \cellcolor{MYCOLOR0}  $k(\x,\z)\colon \mathcal{X}\times \mathcal{X}\to\mathbb{R}$ & \multicolumn{3}{l|}{kernel function.}\\ 
          \cellcolor{MYCOLOR0}  $\K$ & \multicolumn{3}{l|}{$m_t\times m_t$ kernel matrix.}\\ 
           \cellcolor{MYCOLOR0}
          ${\bf k}_{x}=[k(\x,\x_1),\ldots,k(\x,\x_{m_t})]^{\intercal}$ &  \multicolumn{3}{l|}{$m_t \times 1$ vector.}\\
          \hline
\end{tabular}
\label{tab:notation}
\end{table}

\subsection{Acquisition function} 

We consider acquisition functions $A_{t}(\x)\colon \mathcal{X}\to\mathbb{R}$ obtained by the multiplication of a {\em geometry} term $G_t(\x)$ and a {\em diversity} factor $D_t(\x)$, i.e. functions of the form: 
\begin{eqnarray}
\label{AF}
A_{t}(\x)&=&\left[G_t(\x)\right]^{\beta_t}  D_t(\x), 
\end{eqnarray}
where $\beta_t\in[0,1]$ is a positive non-decreasing function of $t$, with $\textstyle\lim\limits_{t\to \infty} \beta_t=1$. 
The function $G_t(\x)$ encodes the geometrical information in ${\bf f}(\x)$, while function $D_t(\x)$ depends on the distribution of the points in the current vector $\X_t$. More specifically, $D_t(\x)$ takes  greater values around empty areas in $\mathcal{X}$, whereas $D_t(\x)$ will be approximately zero close to the nodes and exactly zero at the nodes\footnote{\textcolor{black}{Note that this is the case only for an interpolator (no output-noise assumed) while for a regressor the value of $D_t(\x)$ will just be very small around already placed nodes.}}, i.e., $D_t(\x_i)=0$, for $i=1,\ldots,m_t$ and $\forall t\in \mathbb{N}$. As a consequence, we have  
\begin{eqnarray}
\label{AF_cond2}
A_{t}(\x_i)=0 \quad \forall i, t. 
\end{eqnarray}
{Generally,} since ${\bf f}(\x)$ is { analytically intractable}, the function $G_t(\x)$ can only be derived from information acquired in advance or by considering the approximation $\widehat{{\bf f}}_t(\x)$. The {\em tempering parameter}, $\beta_t$, helps to down-weight the likely less informative estimates of the gradient in the very first iterations. {For instance,} if $\beta_t=0$, we {ignore} $G_t(\x)$ and $A_{t}(\x)= D_t(\x)$, i.e., only the exploration term is considered. Whereas, if $\beta_t=1$, we have $A_{t}(\x)=G_t(\x) D_t(\x)$.


\begin{table}[!t]
	\centering
	\caption{\normalsize {Generic Active \cmmnt{automatic} Emulator.}}
	\vspace{0.2cm}
	    \begin{tabular}{|p{0.95\columnwidth}|}
		\hline
\begin{enumerate}
\item Set $t=0$, select initial points $\X_0=[\x_1,\cdots,\x_{m_0}]$, and ${\bf Y}_0=[{\bf y}_1,\ldots, {\bf y}_{m_0}]$, and maximum number of nodes $M$.
\item While  $m_t <  M$:
\begin{enumerate}
\item Given $\X_t=[\x_1,\cdots,\x_{m_t}]$ and ${\bf Y}_t=[{\bf y}_1,\ldots, {\bf y}_{m_t}]$, build function $\widehat{\f}_t(\x)$.
\item Build the acquisition function $A_t(\x)$ from $\widehat{\f}_t$, and obtain the new input
\begin{equation}
\x_{m_t+1}=\arg\max_{\x\in \mathcal{X}} A_{t}(\x).
\end{equation}
\item Obtain outputs ${\bf y}_{m_t+1}=\f(\x_{m_t+1})$.
\item Update $\X_{t+1} =[\X_t, \x_{m_t+1}]$, ${\Y}_{t+1}=[{\bf Y}_t, {\bf y}_{m_t+1}]$.
\item Set $m_{t+1}=m_t+1$ and $t\leftarrow t +1$.
\end{enumerate}
\item Build the {interpolating} function $\widehat{\f}_t(\x)$. 
\item Return final set of optimal nodes $\{\x_k,\y_k\}_{k=1}^{m_t}$ as a Look-up Table (LUT), as well as the gradient and the predictive variance of the predictive model $\widehat{\f}_t(\x)$.
\end{enumerate}	\\
	\hline
	\end{tabular}		
	\label{alg:GenAE}
\end{table}

\subsection{Specific implementation}

The AE algorithm introduced is completely defined by the choice of the interpolator providing the approximation $\widehat{{\bf f}}_t(\x)$, and the functions $G_t(\x)$, $D_t(\x)$, and $\beta_t$. 
Moreover, the initial set of nodes $\{\x_k,\y_k\}_{k=1}^{m_0}$ and the stopping condition could be considered as additional elements. It is important to note that, in order to choose the interpolating function, we have to take into account the ease of application in high dimensional spaces and the possibility of computing the gradient and other differential geometric measures of $\widehat{{\bf f}}_t$ analytically. 
Different designs of these four elements give rise to different AE techniques. In Section \ref{sec:3}, we provide some specific examples of the choice of $\{\widehat{{\bf f}}_t,G_t,D_t,\beta_t\}$.

\subsection{Parsimonious sequential approach}

It is also important to remark that  the active \cmmnt{automatic} emulation procedure presented in this work is intrinsically a sequential technique. This means that the nodes in $\X_{t-1}$ are always contained in $\X_{t}$, i.e. the locations of previous nodes are not changed. This  solution  minimizes the number of evaluations of the {complex} system $\f$. In this sense, the active \cmmnt{automatic} emulation procedure is a parsimonious sequential technique that { applies}, at each iteration, all previously obtained information about the underlying function $\f$. { Namely, all the previous evaluations of $\f$ are used, and only one additional evaluation of $\f$ is required at each iteration.    

\subsection{Products of the algorithm}
The active emulation procedure proposed in this work is a methodology that delivers: (a) an accurate GP emulator (considering a specific choice of the interpolator) while evaluating the computer code as little as possible,   (b) a final set of nodes  $\{\x_k,\y_k\}_{k=1}^{m_t}$ as a Lookup Table (LUT; other interpolation procedures can be applied using the obtained set of points), and (c) useful statistical information about the model $\f$, such as predictive variance and gradients of the learned function, which can be further used for model inversion and error propagation analyses. 


\section{Active \cmmnt{automatic} Multi-Output Gaussian Process Emulator (AMOGAPE)}\label{sec:3}
An active \cmmnt{automatic} emulator is completely defined by the choice of the {predictive, model $\widehat{\f}({\bf x})$} and the acquisition function $A_t({\bf x})$. In this work, we consider a GP interpolator, as well as the regression formulation~\cite{rasmussen06}, which has been successfully used in remote sensing applications recently~\cite{campsvalls16grsm}.

\subsection{The Gaussian Process Interpolator}

For the sake of simplicity, let us first start considering the GP solution for the scalar output case, i.e., $P=1$. Hence, in this case the vectorial function ${\bf y}={\bf f}({\bf x})$ is a simple function $y=f({\bf x})$, and the matrix 
${\bf Y}_t=[y_{1,1},\ldots,y_{1,m_t}]$, becomes a $1\times m_t$  vector. Given a generic test input $\x$, GPs provide a Gaussian predictive density $p(y|\x)=\mathcal{N}(y|\mu(\x),\sigma^2(\x))$
  with predictive mean $\mu(\x)$ and variance $\sigma^2(\x)$. The predictive mean gives us the interpolating function and  is given by
\begin{eqnarray}
\widehat{f}_t(\x)= \mu_t(\x) ={\bf k}_{x}^{\intercal} {\bf K}^{-1} {\bf Y}_{t}^{\intercal},
\end{eqnarray}
where we defined a kernel function $k(\x,\z)\colon \mathcal{X}\times \mathcal{X}\to\mathbb{R}$, the corresponding kernel matrix
$[{\bf K}]_{ij}:=k(\x_i,\x_j)$ of dimension $m_t\times m_t$ containing all kernel entries, and the kernel vector ${\bf k}_{x}=[k(\x,\x_1),\ldots,k(\x,\x_{m_t})]^{\intercal}$ of dimension $m_t \times 1$. The interpolating function can be simply expressed as a linear combination of $\hat f_t(\x) = {\bf k}_{x}^{\intercal}{\bm \alpha} = \sum_{i=1}^{m_t} \alpha_i k(\x,\x_i)$, 
where the weights ${\bm \alpha}=[\alpha_1,\ldots,\alpha_{m_t}]^{\intercal}$ are ${\bm \alpha}={\bf K}^{-1} {\bf Y}_{t}^{\intercal}$. The GP formulation also provides an expression for the predictive variance
\begin{equation}
 \sigma_t^2(\x) = k(\x,\x)- {\bf k}_{x}^{\intercal}{\bf K}^{-1}{\bf k}_{x}.
\end{equation}
An example is the exponentiated quadratic kernel function, 
\begin{eqnarray}
k(\x,\z)=\exp\left(-\frac{\|\x-\z\|^2}{2\delta^2}\right),
\label{KerGauss}
\end{eqnarray}
where $\|\cdot\|$ is the $\ell_2$-norm, and $\delta>0$ is a positive scalar hyper-parameter.
Note that the norm of the gradient of the interpolating function $\widehat{f}_t$ w.r.t. the input data $\x$ can be easily computed,
\begin{eqnarray}
\texttt{Gr}_t(\x) = \left\|\nabla_x\widehat{f}_t(\x)\right\| = \left\| \sum_{i=1}^{m_t} \alpha_i\nabla_x k(\x,\x_i)\right\|.\label{Geq}  
\end{eqnarray}
The gradient vector of $k(\x,\x_i)$ with $\x=[x_1,\ldots,x_D]^{\intercal}$ and $\x_i=[x_{1,i},\ldots,x_{D,i}]^{\intercal}$, is 
\begin{equation}
\small
\nabla_x k(\x,\x_i)=-\frac{k(\x,\x_i)}{\delta^2}[(x_1-x_{1,i}),\ldots,(x_D-x_{D,i})]^{\intercal},\label{derivada}
\end{equation}
which can be easily computed analytically, and by automatic differentiation software. At this point an intuitive choice of acquisition function of Eq.~\eqref{AF} presents itself. The predictive variance which describes the uncertainty of the GP prediction, and which largely depends on distance to nearby training points, is a natural choice for the diversity term $D_t(\x)= \sigma_t^2(\x)$. Furthermore, since the emulator is differentiable, we can use the gradient as a measure of function variation and choose the geometry term as $G_t(\x)=\texttt{Gr}_t(\x)$.

\subsection{Multi-output GP interpolator}
Several multi-output GP schemes have been proposed with the aim of exploiting the correlation among the output variables~\cite{alvarez2010efficient,alvarez-vector12,hankin2012introducing,alvarez2013linear,Wilson12,Luengo16mlsp}. These models are especially well suited for multitask problems where little data is available or for gap filling, which is not the scenario of this work \cite{alvarez-vector12}. We do not face such problems in our particular remote sensing application since the RTMs provide all vector components when executed in forward mode. 
We adopt a simpler yet highly effective approach, simply treating each output independently. For simplicity, we consider an isotopic case where to each input ${\bf x}_k$ we have $P$ different outputs, $[y_{1,k},...y_{P,k}]^{\intercal}$; see the descrption of isotopic and heterotopic models in \cite{alvarez-vector12}.    
We also define the $p$-th row of the matrix ${\bf Y}_t$ as ${\bf \widetilde{y}}_{p,t}=[y_{p,1},\ldots,y_{p,m_t}],$ with $p=1,\ldots,P$, 
 so that ${\bf Y}_t$ is matrix of dimension $P\times m_t$. Here, for the sake of simplicity, we apply one GP interpolator for each output independently, i.e.,
\begin{gather}
\widehat{{\bf f}}_t(\x)=\left\{
\begin{split}
&\widehat{f}_{1,t}(\x) ={\bf k}_{x,1}^{\intercal} {\bf K}_{1}^{-1} {\bf {\widetilde y}}_{1,t}^{\intercal} \\
& \quad \quad \vdots\\
&\widehat{f}_{P,t}(\x) ={\bf k}_{x,P}^{\intercal} {\bf K}_P^{-1} {\bf {\widetilde y}}_{P,t}^{\intercal}
\end{split}, \right.
\end{gather}
where the vectors ${\bf k}_{x,p}$ have all dimension $m_t\times 1$ and the matrices ${\bf K}_p$ have dimension $m_t\times m_t$. The subindex $p$ in the kernel vector ${\bf k}_{x,p}$ and the kernel matrix ${\bf K}_p$ denotes the dependence to a different hyper-parameter $\delta_p$ (we learn one for each output). More generally, we can consider a different kernel for each output which allows for much model flexibility. Hence,  for each output, we have a different variance 
\begin{equation}
 \sigma_{p,t}^2(\x) = k_p(\x,\x)- {\bf k}_{x,p}^{\intercal}{\bf K}_p^{-1}{\bf k}_{x,p}.
\end{equation}
Similarly, we have one gradient norm for each interpolating function $\texttt{Gr}_{p,t}(\x)$.
It is important to note here that any multi-output GP framework will fit in the AMOGAPE method as long as it provides a differentiable predictive variance and gradient, which for most commonly used kernels is the case.

\subsection{The acquisition function}

Note that $\sigma_{p}^2(\x_i)=0$ for all $i=1,\ldots, m_t$ and all $p$, and each $\sigma_{p}^2(\x)$ depends on the distance among the support points $\x_t$, the chosen kernel function $k$, and the value of the corresponding hyper-parameter $\delta_p$. For this reason, it is reasonable to consider as diversity term the following function that combines them all:
\begin{equation}
D_t(\x):= \sigma_{1,t}^2(\x) \odot \sigma_{2,t}^2(\x) \odot \sigma_{3,t}^2(\x)  ...\odot \sigma_{P,t}^2(\x), \end{equation}
where $\odot$ represents a generic mathematical operation such as sum ($+$) or multiplication ($\times$). 
We wish to use the geometric information term to sample where the norm of the gradient is high and thus define similarly
\begin{equation}
G_t(\x):=\texttt{Gr}_{1,t}(\x) \odot \texttt{Gr}_{2,t}(\x) \odot\texttt{Gr}_{3,t}(\x) ...\odot \texttt{Gr}_{P,t}(\x).
\end{equation}
The intuition behind this choice is that wavy regions of $\f$ (estimated by $\widehat{\f}_t$) require more support points than flat regions. In \ref{appendix} we demonstrate the importance of the gradient term using the simple example of a piecewise-constant interpolator. 
As previously mentioned, we define the acquisition function as
\begin{equation}\label{eq:Ax}
A_{t}(\x)=\left[G_t(\x)\right]^{\beta_t}  D_t(\x).
\end{equation}
Table \ref{tab:acq_now} shows several combinations that generate different acquisition functions according to the choice of the operator $\odot$. 

{\paragraph{Optimization approaches.} 

The maximization of Eq. \eqref{eq:Ax} can be performed by using different optimization algorithms, e.g., gradient ascent or simulated annealing. As can be seen in the simple 1-D example of Fig.~\ref{fig:example}, the acquisition function has many local optima. Thus, while it is useful to have access to the gradient of Eq.~\ref{eq:Ax}, we find that it is important to incorporate stochasticity in the optimization. This can be done, for example by performing a number of random searches and then performing gradient ascent, initialized at the best candidate point.

{\paragraph{Tempering of the geometric information.} 
The parameter $\beta_t\in[0,1]$ indicates how the acquisition function should ``trust'' the provided geometric information and must be an non-decreasing function of $t$. 
Indeed, recall that the geometric information is given by analyzing the interpolating function $\widehat{\f}$, instead of the complex system $\f$, since it is analytically intractable. Clearly, $\beta_t$ must be  an increasing function with respect to $t$, since at each iteration the interpolating function $\widehat{\f}$ is improved and becomes step-by-step more reliable. One possible choice is $\beta_t=1-\exp(-\gamma t)$, where $\gamma\geq 0$ is a positive scalar established by the user or, alternatively, $\beta_t=1-\frac{1}{t}$, for instance.}

\begin{table}[t!]
\centering
\caption{\normalsize Acquisition functions for a multi-output emulator and their shorthand notation used in Section \ref{sec:4}.}
\vspace{-0.0cm}
\label{tab:acq_now}
{\renewcommand{\arraystretch}{1.7}%
\begin{tabular}{|l|l|}
\hline
$\hspace{2cm} A_t(\x)$ & Shorthand \\ \hline
$\sum_{p=1}^P \sigma_{p,t}^2(\x)$ & $\Sigma$D     \\ 
$\prod_{p=1}^P \sigma_{p,t}^2(\x)$ & $\Pi$D      \\ 
$\sum_{p=1}^P \sigma_{p,t}^2(\x)\sum_{p=1}^P \texttt{Gr}_{p,t}(\x)$  & $\Sigma$D$\times$$\Sigma$G \\ 
$\sum_{p=1}^P \sigma_{p,t}^2(\x)\prod_{p=1}^P \texttt{Gr}_{p,t}(\x)$  &  $\Sigma$D$\times$$\Pi$G \\ 
$\prod_{p=1}^P \sigma_{p,t}^2(\x)\sum_{p=1}^P \texttt{Gr}_{p,t}(\x)$  &   $\Pi$D$\times$$\Sigma$G \\ 
$\prod_{p=1}^P \sigma_{p,t}^2(\x)\prod_{p=1}^P \texttt{Gr}_{p,t}(\x)$  &  $\Pi$D$\times$$\Pi$G \\ 
\hline
\end{tabular}}
\end{table}

\subsection{From interpolation to regression}
So far we have described the emulation as an interpolation problem since RTMs are deterministic models: running the code multiple times will always return identical answers. Hence, we have assumed an observation equation of type $y=f(\x)$\footnote{In this section, we assumed only one output in the equation, just for the sake of simplicity. Clearly, the same considerations are valid for the multi-output case.}. However, in some cases, it is preferable to consider an observation equation of type $y=f(\x)+\epsilon$ where $\epsilon\sim \mathcal{N}(0,\upsilon^2)$ represents a Gaussian noise perturbation with zero mean and variance $\upsilon^2$. There are three main reasons, both theoretical and practical, for considering noisy outputs: (a) the system to emulate actually contains stochastic elements (i.e., it is not a completely deterministic system), (b) to increase the prediction power of the emulator function $\widehat{f}_t(\x)$ providing more flexibility to the GP model, and (c) in order to avoid numerical problems, increasing also the stability of the computation. This last point is due to the fact that the noise variance $\nu^2$ plays the role of a regularization term which is added to the diagonal of the kernel matrix (also called a \textit{nugget} in kriging literature). Indeed, when noisy outputs are assumed and by denoting the $m_t\times m_t$ identity matrix as ${\bf I}$, then the GP regression equations become
\begin{eqnarray}
\widehat{f}_t(\x)&=& {\bf k}_{x}^{\intercal} ({\bf K}+\upsilon^2 {\bf I})^{-1} {\bf Y}_{t}^{\intercal}, \\
 \sigma_t^2(\x) &=& \upsilon^2 +  k(\x,\x)- {\bf k}_{x}^{\intercal}({\bf K}+\upsilon^2 {\bf I})^{-1}{\bf k}_{x}.
\end{eqnarray}
Note that, if we set again $D_t(x)=\sigma_t^2(\x)$ (with $\sigma_t^2(\x)$ defined above), then $A_t(\x)$ does not fulfil Eq.~\eqref{AF_cond2}. However, $A_t(x)$ still takes greater values far from nodes $\x_i$, and smaller values close to points $\x_i$.
If the application strictly requires that the condition in Eq. \eqref{AF_cond2} must be satisfied, then we can simply define $D_t(x)=k(\x,\x)-{\bf k}_{x}^{\intercal}{\bf K}^{-1}{\bf k}_{x}$, i.e., $\sigma^2(\x)$ without the noise term. With this definition, we have again $A_t(\x_i)=0$. 
This means that the noise term is only used in the GP equations and not for the construction of the acquisition function. Finally, note that, if a regressor is applied instead of an interpolator, then two hyperparameters must be tuned, $\delta$ and $\upsilon$, instead of just only $\delta$, assuming the kernel in Eq. \eqref{KerGauss}.
 The user might also wish to decide a value of $\upsilon^2$  in advance instead of learning it, using it as a regularization term in order to guarantee the numerical stability of the method. 
Hyperparameter tuning can be performed with standard Cross Validation (CV) procedures, or maximizing the marginal likelihood function by gradient ascent or other optimization techniques \cite{omcmc,kirkpatrick83}. In the interpolation case or when $\upsilon^2$ is decided in advance by the user, another interesting approach is to find the maximum value of bandwidth  $\delta$ which still allows the numeric inversion of the matrix ${\bf K}$ (imposing a upper bound for its condition number).

\section{Experimental results}\label{sec:4}

This section presents experimental results of the our AE framework in synthetic and real (Earth-observation) systems. The AMOGAPE\footnote{Code available at https://github.com/dhsvendsen/AMOGAPE} method is compared to standard algorithms in the literature, namely random exploration/sampling and most notably Sobol's sampling~\cite{Bratley198888} and the Latin Hypercube Sampling (LHS) method~\cite{McKay79}. Algorithms are compared in terms of accuracy and convergence rates in problems of different input and output dimensionality. The real experiments involve a widely used code that models the relation between vegetation parameters and the corresponding reflectance signal.

\subsection{Toy Experiment 1: Example of unidimensional multi-output emulation}  

We consider a multi-output toy example with scalar inputs $x\in{\mathbb R}$ where we can easily compare the achieved approximation $\widehat{{\bf f}}_t(x)=[\widehat{f}_1(x),\widehat{f}_2(x)]$ with the underlying function ${\bf f}(x)=[f_1(x),f_2(x)]$. 
In this way, we can exactly check the true accuracy of the obtained approximation using different schemes. For the sake of simplicity, we  consider the following multi-output mapping
\begin{equation}
\label{TRUE_f}
{\bf f}(x)=[\log(x),0.5\log(3x)], \quad x\in(0,10],
\end{equation}
then $D=1$ and $P=2$ (two outputs). Even in this simple scenario, the procedure used for selecting new points is relevant. We start with $m_0=4$ support points, $\X_0=[0.1, 3.4, 6.7, 10]$, apply an independent GP per output, and for AMOGAPE we use the acquisition function denoted as $\Pi D \times \Pi G$ in Table \ref{tab:acq_now} with the tempering function $\beta_t=1-\frac{1}{t}$. We also set $\upsilon^2=0.02$ as a regularization term, in order to avoid numerical issues.

\subsubsection{Comparison among sequential methods}
 It is important to remark that all the active \cmmnt{automatic} emulators presented in this work are intrinsically sequential techniques. This means that the nodes in $\X_{t-1}$ are always contained in $\X_{t}$, i.e., the previous configuration of points is always kept. Therefore, for a fair comparison we have to consider other sequential algorithms.
We add to $\X_t$ sequentially  $20$ additional points, using different sampling strategies: AMOGAPE, uniform points randomly generated in $(0,10]$, a sequential Sobol sequence,  and a sequential version of the Latin Hypercube Sampling procedure (Seq-LHS).
Seq-LHS simply generates $20$ nodes following the LHS procedure and then adds one to ${\bf X}_t$ at each iteration (without replacement).  
Note that, at each run, the results can vary even for the deterministic procedure due to the optimization of the hyperparameters. We use simulated annealing, which is a stochastic optimization technique~\cite{kirkpatrick83,omcmc}, both for hyperparameter and acquisition function optimization. We average all the results over $500$ independent runs.
For model comparison, we compute the root mean square error (RMSE) between $\widehat{{\bf f}}_t(x)$ and ${\bf f}(x)$ at each iteration, and show the evolution of the (averaged) RMSE versus the number of support points $m_t$ (that is $m_t=t+m_0$) in Figure \ref{FigAMOGAPE}. We can observe that the AMOGAPE scheme outperforms the other methods, providing the smallest RMSEs between ${\bf f}(x)$ and $\widehat{{\bf f}}_t(x)$.

 
\begin{figure}[t!]
\centering

\includegraphics[width=0.75\textwidth]{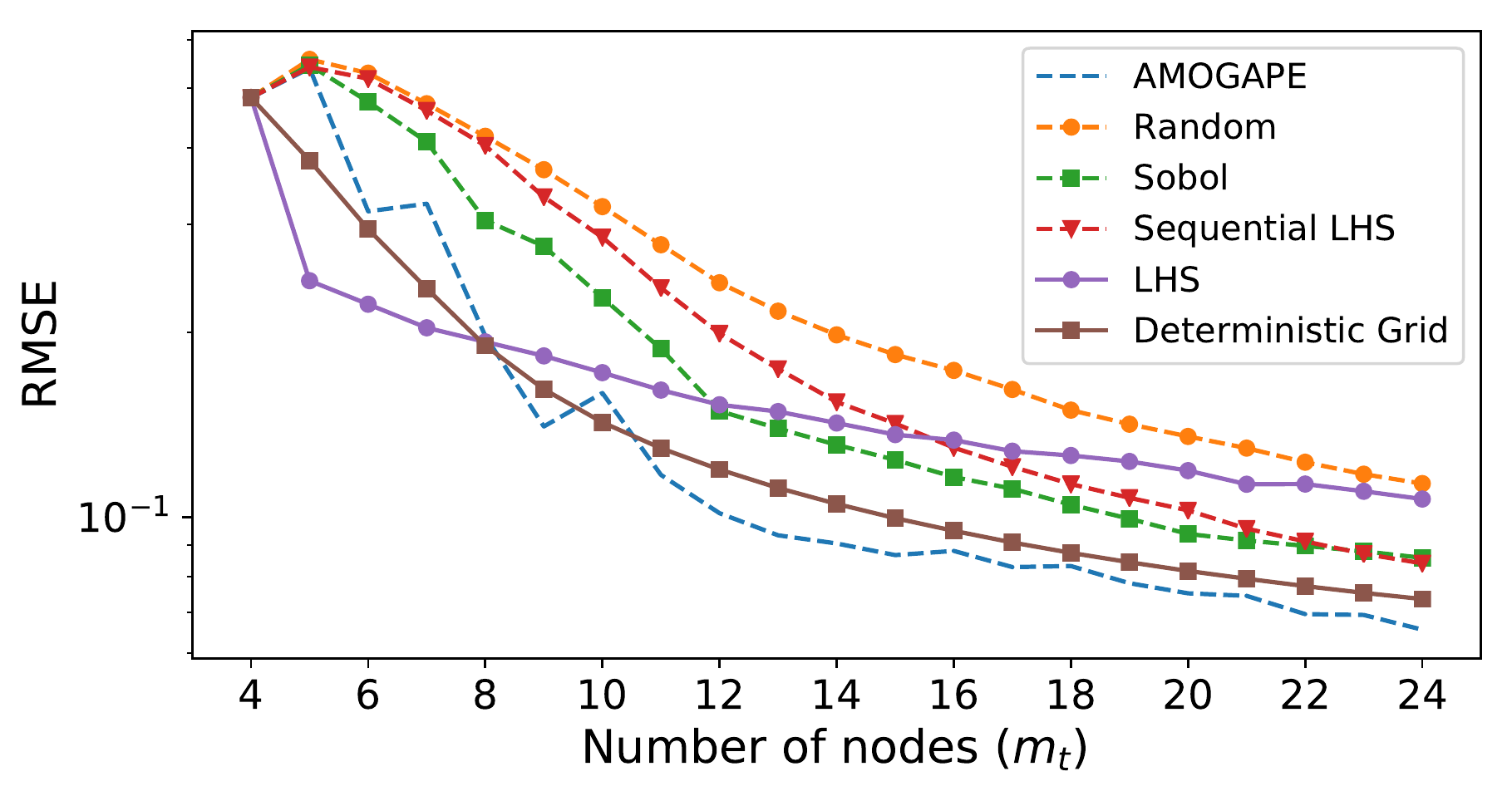}
\vspace{-0.25cm}
\caption{RMSE (in log-scale) between ${\bf f}(x)$ and $\widehat{{\bf f}}_t(x)$ versus the number of nodes $m_t$, that is $m_t=t+4$ in this example ($D=1$ and $P=2$). Sequential methods, which are more comparable as they utilize $m_T$ evaluations of ${\bf f}(x)$, are shown with dashed lines. The comparison with two non-sequential methods, using $\sum_{t=1}^T m_t=\frac{m_T^2+m_T}{2}$ evaluations of ${\bf f}(x)$, are shown with solid lines. 
}
\label{FigAMOGAPE}
\end{figure}

\subsubsection{Comparison with non-sequential methods}
In order to provide an exhaustive numerical analysis we also compare AMOGAPE with non-sequential techniques where the input matrix $\X_{t}$ can be completely different from $\X_{t-1}$ (whereas, in AMOGAPE,  the nodes in $\X_{t-1}$ are all always contained in $\X_{t}$). This approach would not be used in practice, but serves as an interesting comparison of AMOGAPE with one-shot space-filling algorithms.
More specifically, we consider:
\begin{itemize}
\item Deterministic grid: at each step, we consider an equal-spaced set of points (deterministically chosen). Thus, at each step, all the points in the previous timestep $\X_{t-1}$ are not considered but replaced by new nodes. 
\item Standard LHS: also in this case, at each iteration all the previous points are changed.
\end{itemize}
Clearly, these two schemes evaluate the underlying function in $m_t$ new nodes at each iteration and are therefore more costly than AMOGAPE. The total number of evaluations of ${\bf f}(x)$ for AMOGAPE is $m_T$ whereas, for the non-sequential schemes above is $\sum_{t=1}^T m_t=(m_T^2+m_T)/{2}$.
However, even in this unfair comparison for our method, Figure \ref{FigAMOGAPE} shows that AMOGAPE is able to provide the smallest error when more than $12$ new points are incorporated. This illustrates that the gradient term encoded in the AMOGAPE adds useful information to the active learning scheme.

\subsection{Toy Experiment 2: Example of bidimensional multi-output emulation}\label{seq:toy2}  

In this section, we extend the previous example to consider multi-input and multi-output problems, i.e. $D=P=2$. More specifically, we consider
\begin{equation}
\label{TRUE_f_2}
{\bf f}({\bf x})=[\log({|{\bf x}|}),0.5\log(3{|{\bf x}|})], \quad {\bf x}\in(0,10]\times (0,10].
\end{equation}
We start with $m_0=25$ starting nodes in the input matrix, $\X_0=[\x_1,\dots,\x_{m_0=25}]$ where $\x_i=[x_{i,1},x_{i,2}]^{\intercal}$, with $i=1,\ldots, 25$, distributed as shown in Figure \ref{FigAMOGAPE_2}(a) with black circles.
In order to evaluate the approximation RMSE obtained with the emulators, we consider a thin grid in the square $(0,10]\times (0,10]$ (with step $0.3$). The starting nodes in the input matrix, $\X_0$ (black points), and the thin test grid (green dots) are shown in Fig. \ref{FigAMOGAPE_2}(a). 
We apply again one independent GP for each output and, as in the previous example. For AMOGAPE, we apply the acquisition function denoted as $\Pi D \times \Pi G$ in Table \ref{tab:acq_now} and we use again the tempering function $\beta_t=1-\frac{1}{t}$ and set $\upsilon^2=0.02$ as a regularization term, only for avoiding numerical issues. 
We compare different sampling strategies : AMOGAPE, a sequential Sobol sequence, and sequential LHS. 
We add $30$  additional points to $\X_0$ in the first two sequential approaches. In LHS all the previous points change at each iteration.
The results (averaged over 500 independent runs) are shown in Fig.~\ref{FigAMOGAPE_2}(b), which show a considerable gain in accuracy and convergence rates by the presented algorithms.\\
The distributions in input-space of the final 55 nodes - 25 on a grid and 30 subsequently chosen with a sampling algorithm - are shown as a Kernel Density Estimation (KDE) plot for each of the methods in Figure.~\ref{fig:kde}. 
 We can observe that AMOGAPE adds points in the border and in a left-bottom corner where the gradient is comparatively high. It makes sense that these points are deemed the most useful since  the initial 25 nodes points are well-located. The sampling method using the Sobol algorithm and the sequential LHS algorithms incorporate new points that fill out the input-space but do not pay particular attention to the behaviour of the underlying function.

\begin{figure}[t!]
\centering
\centerline{
\subfigure[]{\includegraphics[width=0.5\textwidth]{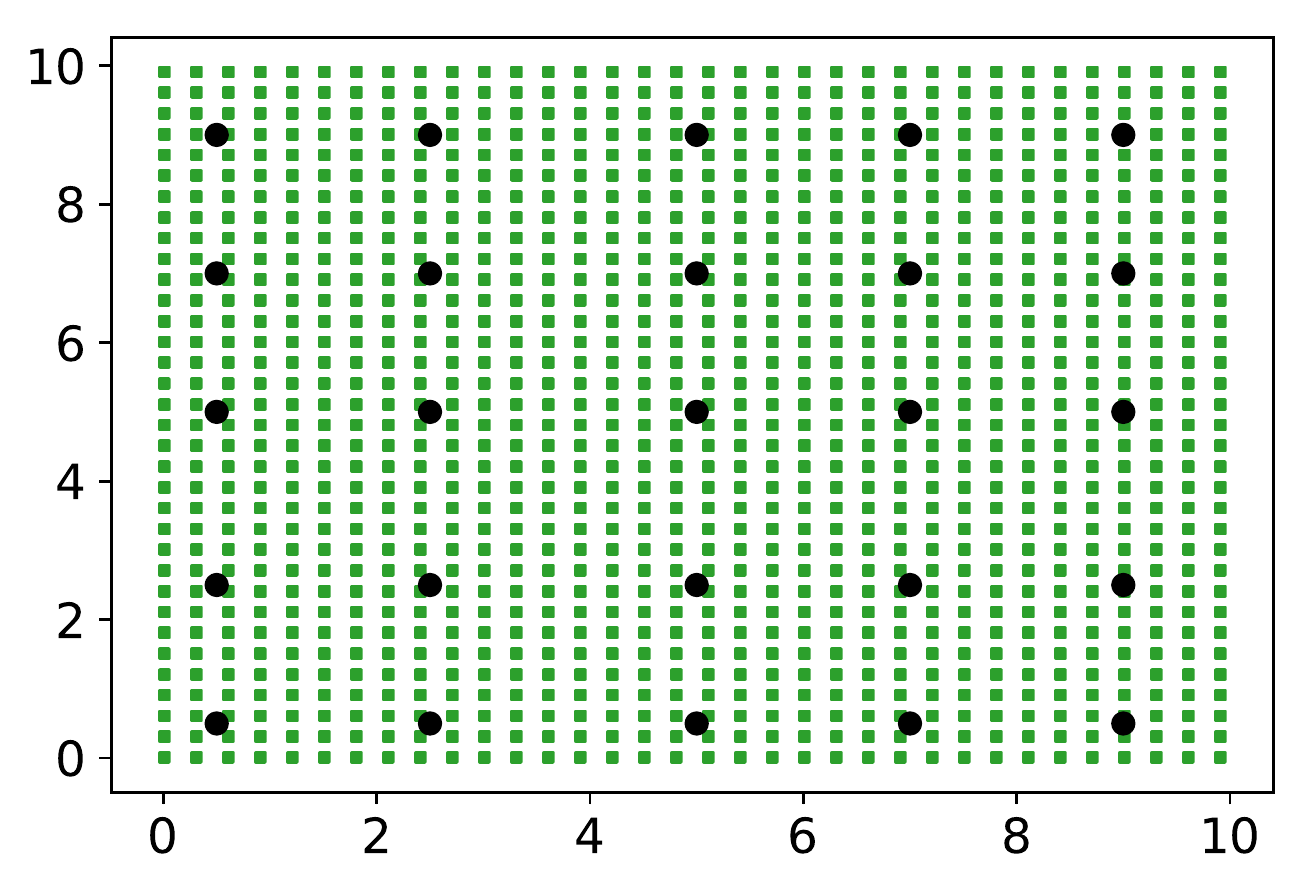}}
\subfigure[]{\includegraphics[width=0.5\textwidth]{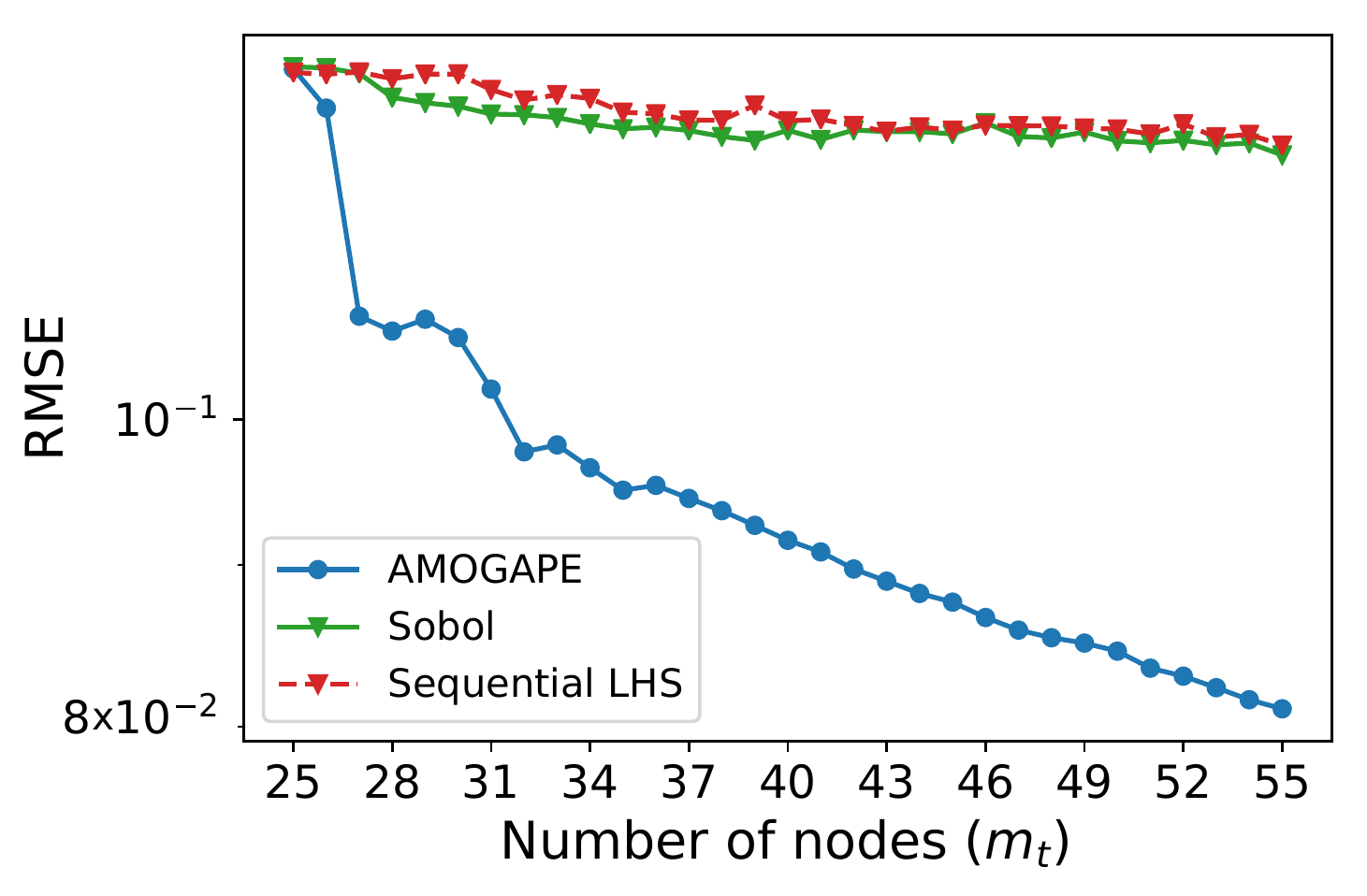}}
}
\vspace{-0.25cm}
\caption{{\bf (a)} The starting points in the input matrix $\X_0$ are shown with black points, whereas the points in the thin test grid are depicted with green squares. {\bf (b)} RMSE (in log-scale) between ${\bf f}(\x)$ and $\widehat{{\bf f}}_t(\x)$ versus the number of the number of support points $m_t$, that is $m_t=.....$ in this example ($D=2$ and $P=2$). Note that the number of initial points is $m_0=25$.
}
\label{FigAMOGAPE_2}
\end{figure}

\begin{figure}[h!]
\centering
\centerline{
\includegraphics[width=1\textwidth]{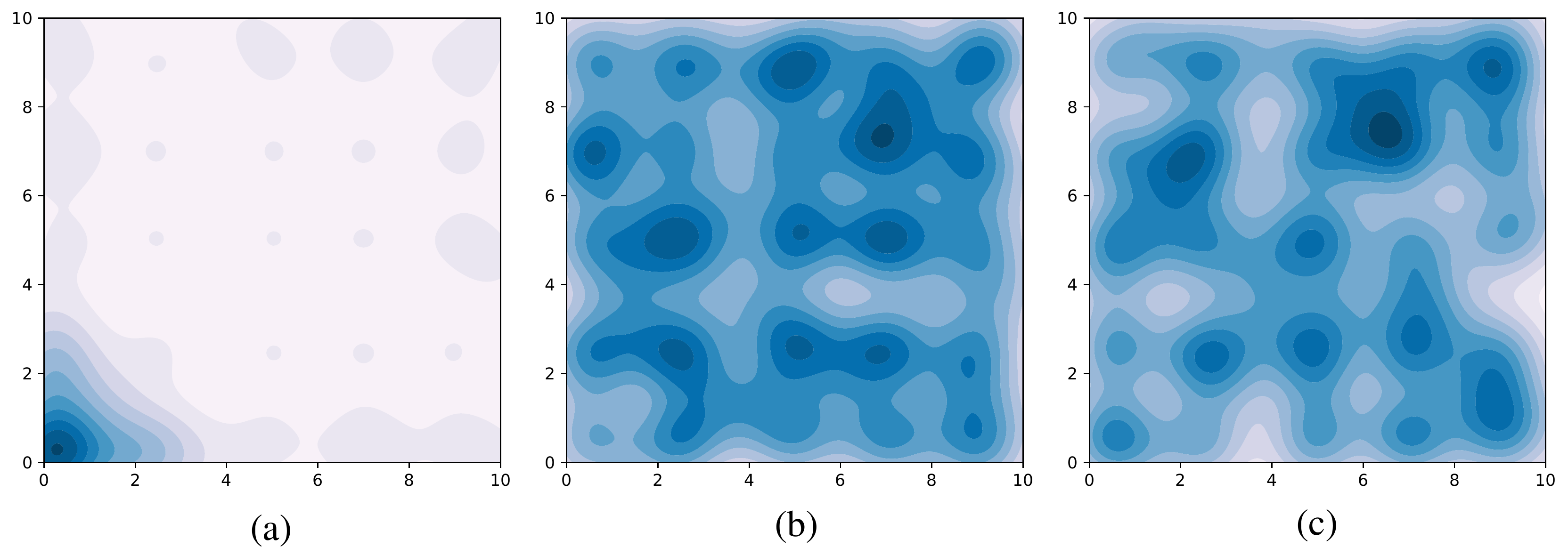}
}
\vspace{-0.25cm}
\caption{Kernel Density Estimation plot of the final configuration of points in the emulation of Eq.~\eqref{TRUE_f_2}, showing the {\bf (a)} AMOGAPE, {\bf (b)} Sobol and {\bf (c)} Sequential LHS algorithms respectively.
}
\label{fig:kde}
\end{figure}

\subsection{Application to remote sensing: Emulating a radiative transfer model} \label{NumSect_PROSAIL}
Our method is assessed for the emulation of the leaf-canopy PROSAIL RTM, which is the most widely used RTM over the last two decades in remote sensing studies \cite{jacquemoud2009prospect+}. PROSAIL simulates reflectance as a function of:
\begin{itemize}
    \item[1.]\emph{Leaf optical properties}, given by the mesophyll structural parameter (N), leaf chlorophyll (Chl), dry matter (Cm), water (Cw), carotenoid (Car) and brown pigment (Cbr) contents. 
    \item[2.]\emph{Canopy level characteristics}, determined by leaf area index (LAI), the average leaf angle inclination (ALA) and the hot-spot parameter (Hotspot). System geometry is described by the solar zenith angle ($\theta_s$), view zenith angle ($\theta_\nu$), and the relative azimuth angle between both angles ($\Delta \Theta$).
\end{itemize}
We consider PROSAIL for simulating \href{https://www.nasa.gov/content/landsat-8-instruments}{Landsat-8} spectra, 
a satellite sensor widely used for land cover applications in general and vegetation monitoring in particular. Therefore, the generated, eventually optimized, look-up tables are used for inversion and thus retrieve vegetation parameters with the Landsat-8 satellite imagery. This leaves us with an output-dimension of $P=9$ for our problem, i.e. the number of spectral bands of the satellite. Now, depending on the parameters of interest the input dimensionality $D$ may vary.

\subsubsection{Sampling a 2-dimensional space for PROSAIL emulation}
In this experiment, we chose the most important variables at leaf and canopy-level respectively, namely Chl and LAI, and kept the rest fixed. Table \ref{tab:profixed} shows the values for the remaining parameters which are set for simulation of rice crops \cite{campostaberner2016b}.
When generating look-up tables with RTMs it is common practice to use expert knowledge to determine distributions over the biophysical parameters which constitute the RTM input \cite{weisss2toolbox}. The desired amount of samples are then drawn, and the model is evaluated in each of these points. A commonly used distribution is the truncated Gaussian $\mathcal{N}_{\mathcal{T}}(\x | \mu , \sigma , \min , \max)$. Indeed, the truncated Gaussians $\mathcal{N}_{\mathcal{T}}$(Chl$|$45, 30, 20, 90) and $\mathcal{N}_{\mathcal{T}}$(LAI$|$3.5, 4.5, 0, 10) for Chl and LAI, respectively, have proven effective for crop reflectance modeling \cite{campostaberner2016b}. We denote their joint distribution, which has no covariance between the variables, as $\mathcal{N}_{\mathcal{T}}$(Chl,LAI).

\begin{table}[h!]
\centering
\caption{Characteristics of the simulation used in the PROSAIL model.\label{tab:profixed}}
\vspace{0.2cm}
\begin{tabular}{|c|c|c|c|c|c|}
\hline
\multirow{2}{*}{\emph{Leaf level}}   & N & \multicolumn{1}{c|}{Cm} & \multicolumn{1}{c|}{Cw} &  Car & Cbr  \\ \cline{2-6} 
                              & 1.5 & \multicolumn{1}{c|}{ 0.01 $\upmu$g/cm$^2$ } & \multicolumn{1}{c|}{0.01 $\upmu$g/cm$^2$} & 8 g/cm$^2$ & 0   \\ \cline{1-6} 
\multirow{2}{*}{\emph{Canopy level}} & ALA & Hotspot & $\theta_s$ & $\theta_\nu$ & $\Delta\Theta$ \\ \cline{2-6} 
                              &  Spherical & 0.01 & 30$^\circ$ & 10$^\circ$ & 0  \\ \hline
\end{tabular}
\end{table}

In summary, we are emulating a function f$(\x)$ where $\x$ = [Chl, LAI] mapping from an input space of dimension $D=2$ to the output space of dimension $P=9$. The search space is restricted to physically meaningful values of Chl $\in[20; \, 90]$ $\mu$g/cm$^2$ and LAI $\in[0;\, 10]$. 
In order to gain insight into the relative importance of the Diversity and Geometric terms, an array of different acquisition functions shown in Table \ref{tab:acq_now} are applied. The AMOGAPE sampling schemes are compared with sampling randomly from $\mathcal{N}_{\mathcal{T}}$(Chl, LAI). This distribution encodes knowledge about the physically feasible region to sample in, which is also encoded in the AMOGAPE, simply by multiplying the truncated density function $\psi$(Chl, LAI) onto the acquisition functions.
We set $\beta_t = 1 \,\, \forall \,\, t$ in order to simplify the experiments.

Evaluation of which sampling method leads to the best emulator is done by computing the test approximation error on a test-set of 5000 points, sampled from the above-mentioned truncated Gaussian distributions. We initialize with $30$ points drawn from $\mathcal{N}_{\mathcal{T}}$(Chl,LAI). The multi-output RMSE for the $M=5000$ test points over the $P=9$ single-output GP emulators is computed as follows
\begin{equation}
    \mbox{RMSE} = \sqrt{  \frac{1}{M} \sum_{i=1}^{M}  \frac{1}{P} \sum_{p=1}^{P}  (  y_{p,i} - \widehat{y}_{p,i} )^2 }.
\end{equation}
The results are averaged over $15$ runs. In order to speed up the experiment, hyperparameter and acquisition function optimization are performed through an initial random search of $10^D$ points, followed by gradient ascent.
Results are shown in Fig.~\ref{fig:2dresults}. We see that it is possible to perform better using the AMOGAPE approach on our test-set than by sampling randomly from $\mathcal{N}_{\mathcal{T}}$(Chl,LAI). 
It is interesting to note that methods using $\Sigma$D$\times$$\Sigma$G and $\Sigma$D perform similarly, implying that the $\Sigma$D term is governing the acquisition function. Similarly, methods using $\Pi$D and $\Pi$D$\times$$\Sigma$G perform equally well, showing that $\Pi$D is the most influential term. 
The acquisition function $\Sigma$D$\times$$\Pi$G, which penalizes a zero-gradient in any of the output dimension, relies too much on geometric information and performs the worst. It seems that the information source which is included in product form governs $A(\x)$.
It seems that the $\Pi$D$\times$$\Pi$G method manages to strike a balance between the two sources of information. All in all, the best performing methods are $\Sigma$D and $\Sigma$D$\times$$\Sigma$G. This hints at the idea that the product form is too restrictive, i.e. considering a point uninteresting if the predictive variance is close to zero in only one of the output-dimensions.

\begin{figure}[t!]
\centering
\includegraphics[width=0.9\textwidth]{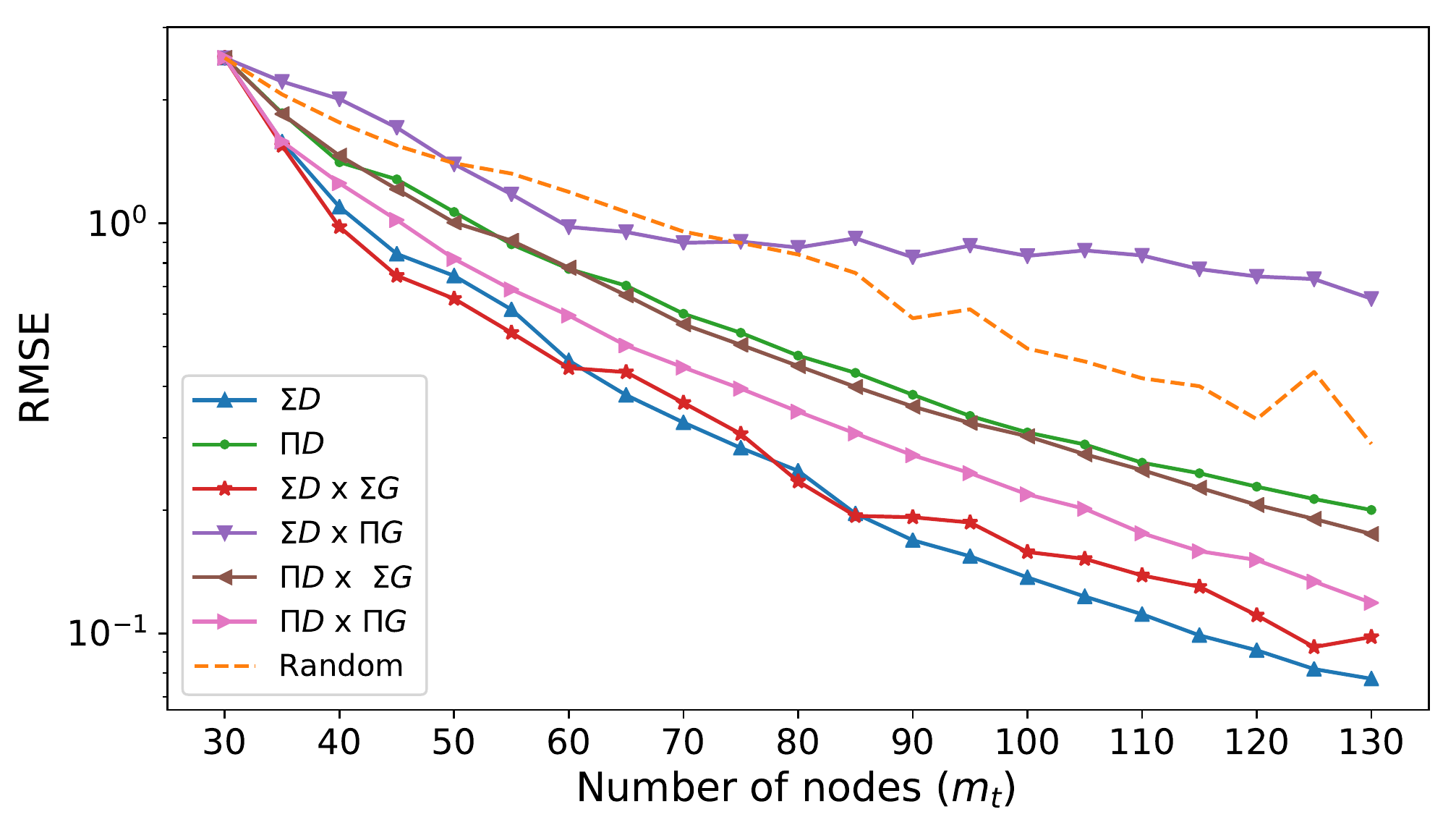}
\caption{Function approximation errors by different acquisition functions, cf. Tab.~\ref{tab:acq_now}, and for different number of selected nodes $m_t$ in a bidimensional PROSAIL problem. Only the best performing acquisition functions are compared here to random sampling.}
 \label{fig:2dresults}
\end{figure}

\subsubsection{Sampling a 3-dimensional space for PROSAIL emulation}
We conduct a similar experiment, including now another crucial biophysical parameter in the search space, namely dry matter content (Cm), which is an important parameter to monitor key properties and processes in vegetation and the wider ecosystem. 
The associated truncated Gaussian used is  $\mathcal{N}_{\mathcal{T}}$(Cm$|$0.005, 0.005, 0.003, 0.011). We use a test set of 50000 points generated from the joint truncated Gaussian $\mathcal{N}_{\mathcal{T}}$(Chl,~LAI,~Cm).

We saw earlier that the acquisition function which performs the best was also the most simple, namely $\Sigma$D. 
The acquisition function $\Pi$D$\times\Pi$G, being formulated only in product form, manages not to be dominated by either term and is interesting because it is very selective: 
It discourages a gradient or predictive variance which is close to zero in any output dimension. For these reasons, along with computational burden, the aforementioned acquisition functions are used for the 3-dimensional experiment. 
The average results after running the experiment $10$ times are shown in Fig.~\ref{fig:3dresults}. Again, we see that the 1) the two variants of AMOGAPE acquisition functions outperform random sampling, 2) that the acquisition functions behave quite similarly, and 3) that simple acquisitions perform as well as more complicated ones. Note however, that using a different tempering function than $\beta_t = 1 \, \forall \, t$ would likely make performances diverge. 

\begin{figure}[h!]
    \centering
    \includegraphics[width=0.9\textwidth]{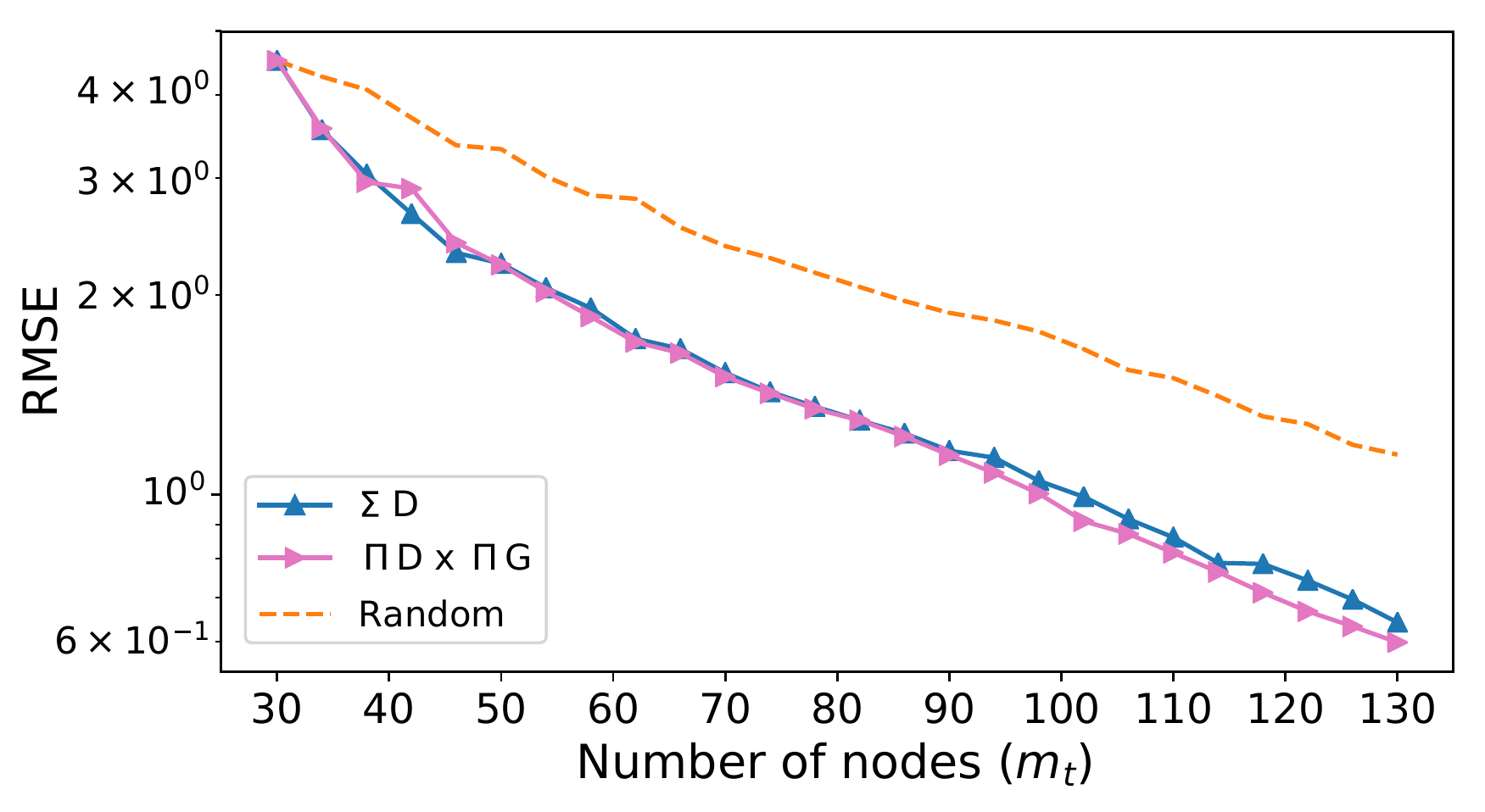}
    \caption{Function approximation errors by different acquisition functions and for different number of selected nodes $m_t$ in the  three-dimensional PROSAIL problem.}
    \label{fig:3dresults}
\end{figure}


\section{Conclusions} \label{sec:5}

We introduced a simple framework for active \cmmnt{automatic} construction of emulators for costly physical models used in Earth observation. 
{The proposed framework does not only provide an effective approximating function, but also a compact LUT and some very useful by-products for practitioners, namely confidence intervals for the estimates and information about the gradients.}

The methodology iteratively incorporates new sample points that meet both diversity and geometry criteria, thus sampling in low-density and more `complex' regions. This is accomplished by building an acquisition function that takes into account the predictive variance and the norm of the gradient of the GP function used for emulation. 
The combination of the geometric and diversity sampling criteria was possible because both the GP predictive variance and the gradient of the GP predictive mean are analytic expressions. 

We illustrated the promising capabilities of the method through emulation of a popular radiative transfer model. Comparison to established methods in the literature illustrated the favourable performance of the proposed methods. The proposed family of criteria for defining the acquisition functions in emulation allows smart sampling of the input space thus leading to compact and expressive look-up-tables, which can be readily used for model inversion in either statistical or numerical frameworks. 
The proposed methodology is very general and modular. Alternative acquisition functions, kernel functions and quality measures adapted to the problem are interesting pathways to explore. In our future work we plan to explore the use of Mat\'ern kernels when function smoothness is not a strict (or even realistic) requirement in a given RTM. Besides, other quality measures other than RMSE could be more interesting for evaluating emulator quality. The information content of the added samples in each iteration by computing maximum differential of entropies in similar ways to the approach in \cite{ruiz2013bayesian}. 

We anticipate adoption of these methods in the Earth sciences and also in unrelated disciplines where process-based models are widely adopted as well, from econometrics to industry or health sciences. Our future work is centered around speeding up other more complex codes, such as the atmosphere MODTRAN model, as well as to extend the framework to deal with dynamic models. The framework introduced here constitutes the first step towards the ambitious goal of large scale active \cmmnt{automatic} statistical models that learn Physics models. 

\section*{Acknowledgments}

This work is supported by the European Research Council (ERC) under the ERC-CoG-2014 SEDAL Consolidator grant (grant agreement 647423).

\bibliography{refs}

\appendix

\section{Importance of a gradient term in the acquisition function}\label{appendix}
Let us consider the problem of approximating the function $y=f({\bf x})$, ${\bf x} \in  \mathcal{D}\subseteq \mathbb{R}^L$, where  $\mathcal{D}=[a_1,b_1]\times [a_2,b_2] \times... \times [a_L,b_L]$.  For the sake of simplicity, we consider one dimensional problems, i.e. $L=1$, with $\mathcal{D}$ bounded $a_i,b_i <\infty$.  Moreover, let us consider a set of nodes $\{{\bf x}_1,{\bf x}_2,...,{\bf x}_{M}\} \in \mathcal{D}$ and the corresponding values of the function $y_m=f({\bf x}_m)$. We use the pairs $\{{\bf x}_m,y_m\}_{m=1}^M$ to perform an interpolation. \\
Given the set of nodes $\{{\bf x}_m\}_{m=1}^M$, we denote a piecewise constant interpolation (PCI) of the function $f({\bf x})$  as 
\begin{eqnarray}
\widehat{y}=\phi({\bf x}|{\bf x}_{1:M})=\phi({\bf x}).
\end{eqnarray}
In order to measure the discrepancy between function and emulator we introduce a cost function $C(f,\phi)=C\left(f({\bf x}),\phi_({\bf x})\right) \geq 0$.
The equality $C(f,\phi)=0$ must hold only if $\phi({\bf x})=f({\bf x})$. Here, we consider the $L_p$ family of cost functions
\begin{eqnarray}
C_p(f,\phi)=||f({\bf x})-\phi({\bf x})||_p=\left(\int_\mathcal{D}|f({\bf x})-\phi({\bf x}) |^p \mbox{d}{\bf x}\right)^{\frac{1}{p}}.
\end{eqnarray}
Note that $C_\infty(f,\phi)=\lim_{p\rightarrow \infty} C_p(f,\phi) = \max\limits_{{\bf x}\in \mathcal{D}} \left|f({\bf x})-\phi({\bf x}) \right|$.

\subsection*{One node ($M=1$), infinity norm cost functions $p=\infty$}
Let us consider $y=f(x)$, $x \in  [a,b]\subseteq \mathbb{R}$. For simplicity, we assume $f(x)$ to be  strictly monotonic, more specifically increasing. 
We consider a piecewise constant approximation with $M=1$ point $x_1$ within $x_1$, i.e.
\begin{gather}
\phi(x)=\left\{
\begin{split}
f(a) \quad x\leq x_1 \\
f(x_1) \quad x>x_1 \\
\end{split}
\right.
\end{gather}
Let us consider the $L_\infty$ distance (i.e., $p=\infty$),
\begin{eqnarray}
C_\infty(x_1)= \max\limits_{ x\in [a,b]} \left|f(x)-\phi_{0}(x) \right|&=&\max\limits_{x\in [a,b]}\left[|f(x_1)-f(a)| , |f(x_1)-f(b)|\right], \nonumber \\
&=&\max\limits_{x\in [a,b]}\left[f(x_1)-f(a), f(b)-f(x_1)\right]
\end{eqnarray}
The problem consists in finding the optimal node $x_1^*$ such that  $x_1^*=\arg\min C_\infty(x_1)$.
\label{Teo1} This optimal point $x_1^*$ will then satisfy the condition
\begin{equation}
\label{ImpEq}
f(x_1^*)-f(a)=f(b)-f(x_1^*).
\end{equation}
\textcolor{black}{This is because $c_a \equiv | f(x_1)-f(a) |$ will decrease as $c_b \equiv | f(x_1) - f(b) |$ increases, and vice versa, due to the monotonicity of $f(x)$. Since we are taking the max between the two, the lowest value that $C_\infty$ can take is the point where they are equal $c_a = c_b$ (see Fig.~\ref{AppFig1}), as any divergence from that would lead to one of the terms being higher.}

Using Eq. \eqref{ImpEq}, since we are assuming that $f$ is monotonic (thus invertible), we can also write
\begin{eqnarray}
f(x_1^*)&=& \frac{f(b)+f(a)}{2},
\end{eqnarray}
and, since we have assumed that $f(x)$ is monotonic, thus invertible, we have
\begin{eqnarray}
 x_1^*=f^{-1}\left(\frac{f(b)+f(a)}{2}\right).
\end{eqnarray}
Figure~\ref{AppFig1} illustrates the above reasoning. Note that, if $f(x)$ is non-linear, $x_1^*\neq \frac{a+b}{2}$ (as a space filling/Latin hypercube strategy might suggest). The expression of  $x_1^*$ is an extended mean, which takes into account information regarding the non-linearity $f(x)$.
\newline
\newline

  \begin{figure*}[htbp]
\centering
\centerline{
\subfigure[]{\includegraphics[width=6.1cm]{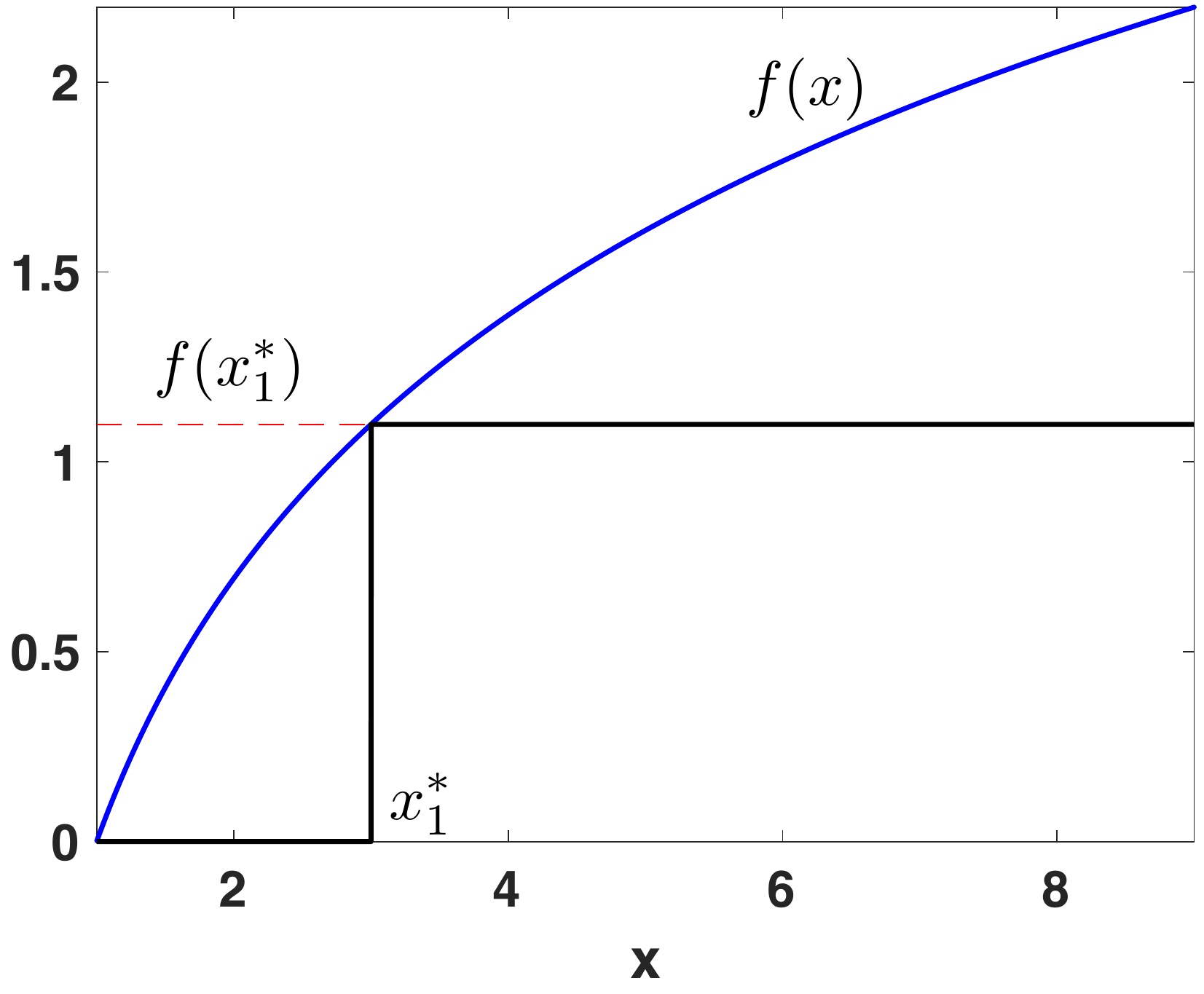}}
\subfigure[]{\includegraphics[width=6.3cm]{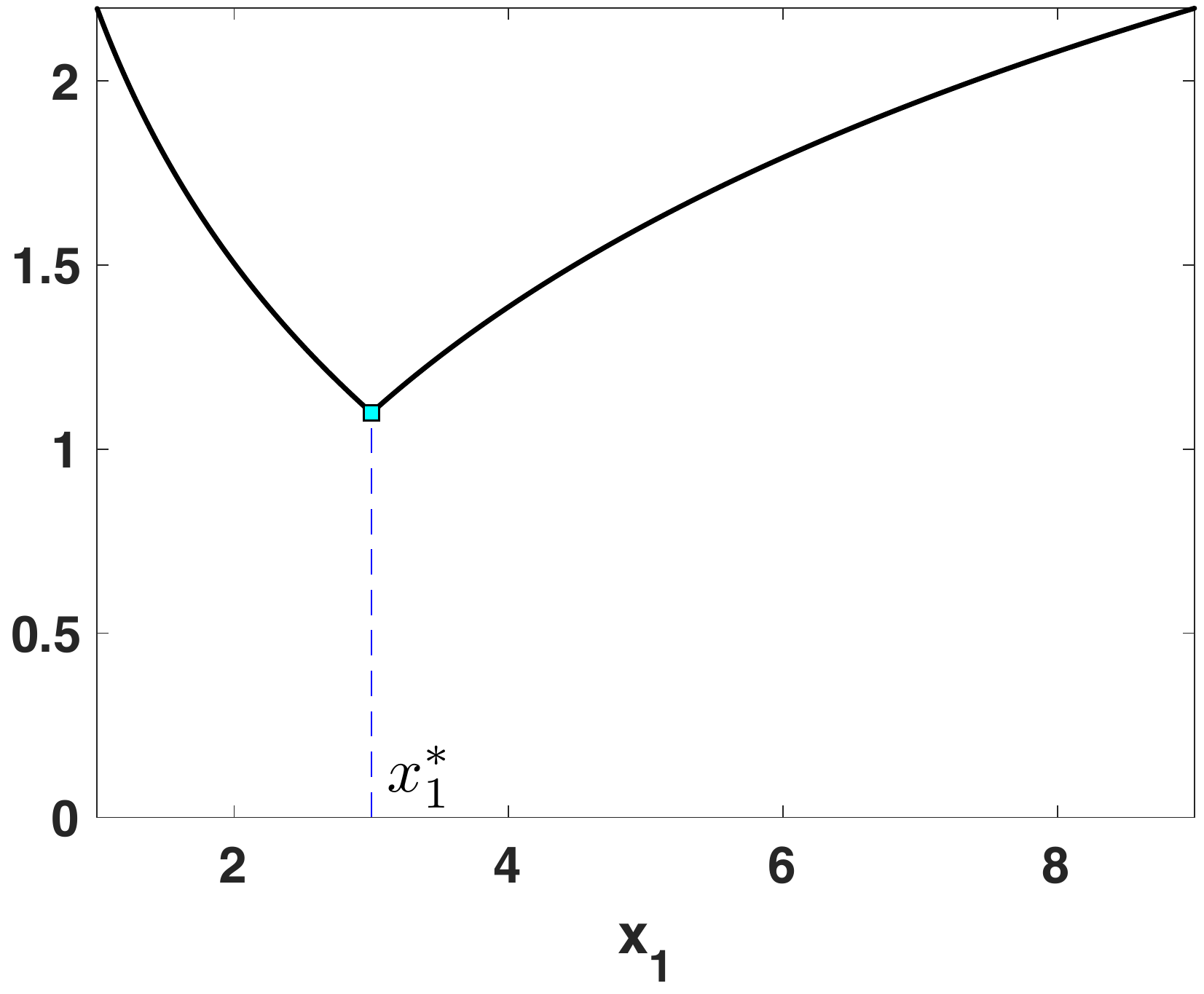}}
}
\centerline{
}
\caption{{\bf (a)} Optimal piecewise constant approximation $\phi(x)$ with $M=1$ node. {\bf (b)} The cost function $C_\infty(x_1)$ and its minimum at $x_1^*$.}
\label{AppFig1}
\end{figure*}

\subsection*{Generic number of nodes $(M>1)$}
Let us assume now that we may place $M$ nodes $x_1,x_2,...,x_M$ in order to achieve an optimal emulator of $f$ with respect to the $C_\infty$ norm. The optimal nodes $x_1^*,x_2^*,\ldots,x_M^*$ will then satisfy the condition
\begin{equation}
\label{ImpEq2}
f(x_1^*)-f(a)=f(x_2^*)-f(x_1^*)=\ldots f(x_{M-1}^*)-f(x_{M-2}^*)=f(b)-f(x_{M-1}^*).
\end{equation}
The point $(x_1^*,x_2^*,\ldots,x_M^*)$ is a minimum for $C_\infty(x_1,x_2,\ldots,x_M)$ and is unique. In order to see this, let us define  $d_1=f(x_1)-f(a)$, $d_2=f(x_2)-f(x_1),\ldots,d_m=f(x_m)-f(x_{m-1}),\ldots,d_M=f(x_M)-f(x_{M-1})$, and $d_{M+1}=f(b)-f(x_{M})$. With this definition we reach the minimum for $C_\infty(x_1,x_2,\ldots,x_M)$ when all distances $\{d_m\}_{m=1}^M$ are equal to  $f(b)-f(a)$ divided by $M+1$:
 \begin{equation}
 \label{eq:optd}
d_1^*=d_2^*=\ldots=d_{M+1}^*=\frac{f(b)-f(a)}{M+1} \equiv c_{\texttt{MIN}}
\end{equation}
This can be seen from the fact that the distances satisfy $d_m \geq 0$ for $m = 1,2,\ldots,M$ due to the monotonicity of $f$, and they sum to a constant
\begin{equation}
    \sum_{m=1}^{M+1}  d_m = f(b) - f(a)
\end{equation}
Thus if one $d_m$ decreases, one or all other have to increase. This implies that any configuration of $x_1,x_2,\ldots,x_M$ resulting in $d_j <  c_{\texttt{MIN}}$ for some $j$, will lead $d_k >  c_{\texttt{MIN}}$ for one or more $k$.
Therefore, the configuration of corresponding to \eqref{eq:optd} is optimal.  \\
Following the above logic, the optimal locations of $M$ nodes, $\{{\bf x}_1,\ldots,{\bf x}_M\}$, can be obtained by:
\begin{enumerate}
\item  Dividing with a uniform grid formed by $M$ points the interval $[f(a),f(b)]$ (image of $[a,b]$),
\begin{equation}
y_m=f(a)+m\frac{f(b)-f(a)}{M+1}, \qquad m=1,\ldots,M, T
\end{equation}
\item Finding the $x_m$ such that $f(x_m)=y_m$, i.e., since we assume that $f(x)$ is invertible,
\begin{equation}
\label{ImpFormula}
x_m=f^{-1}(y_m), \qquad m=1,\ldots,M.
\end{equation}
\end{enumerate}
See Fig.~\ref{AppFig2} for an example with $M=2$ nodes.

 \begin{figure*}[htbp]
\centering
\centerline{
\subfigure[]{\includegraphics[width=6.1cm]{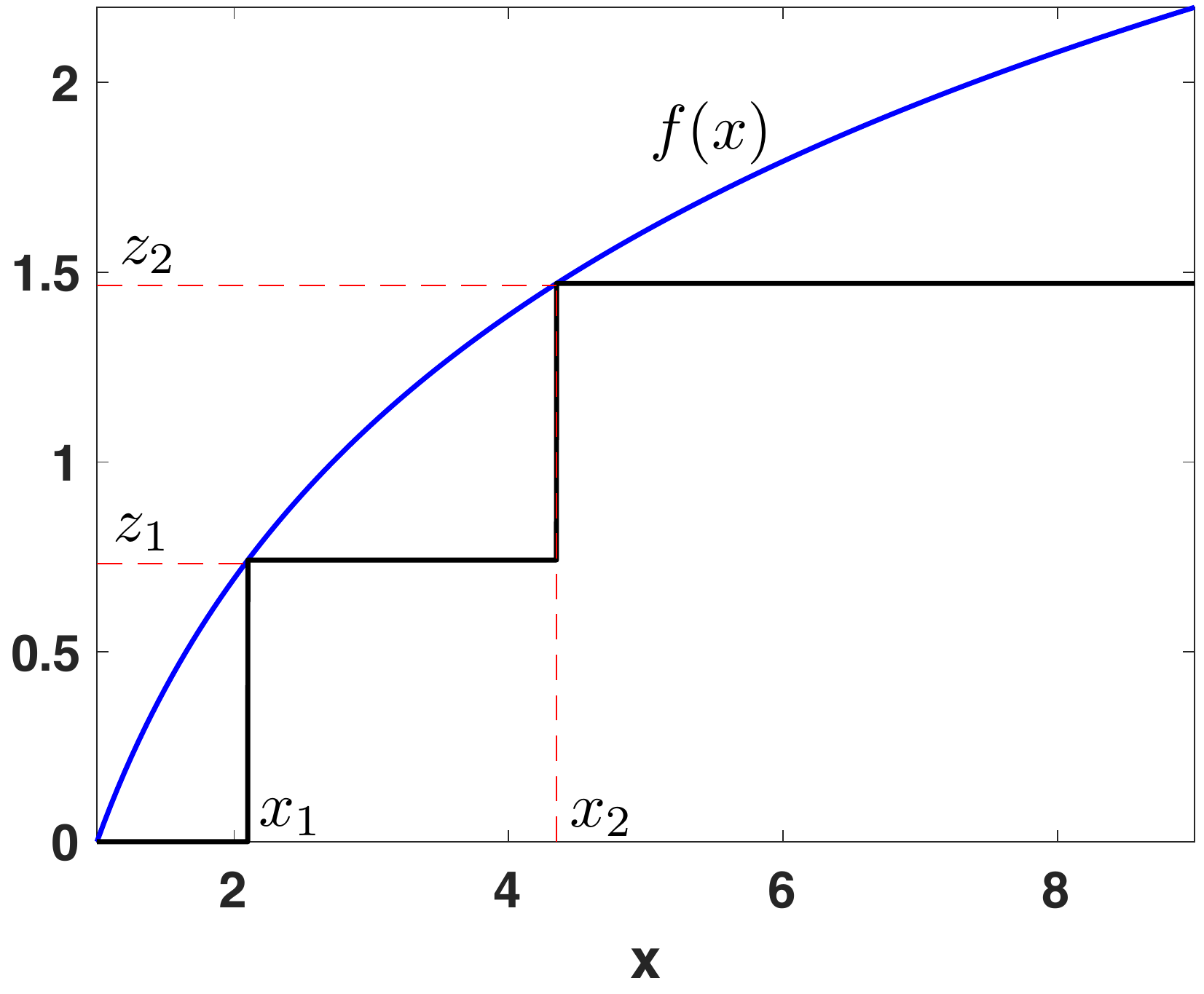}}
\subfigure[]{\includegraphics[width=6.3cm]{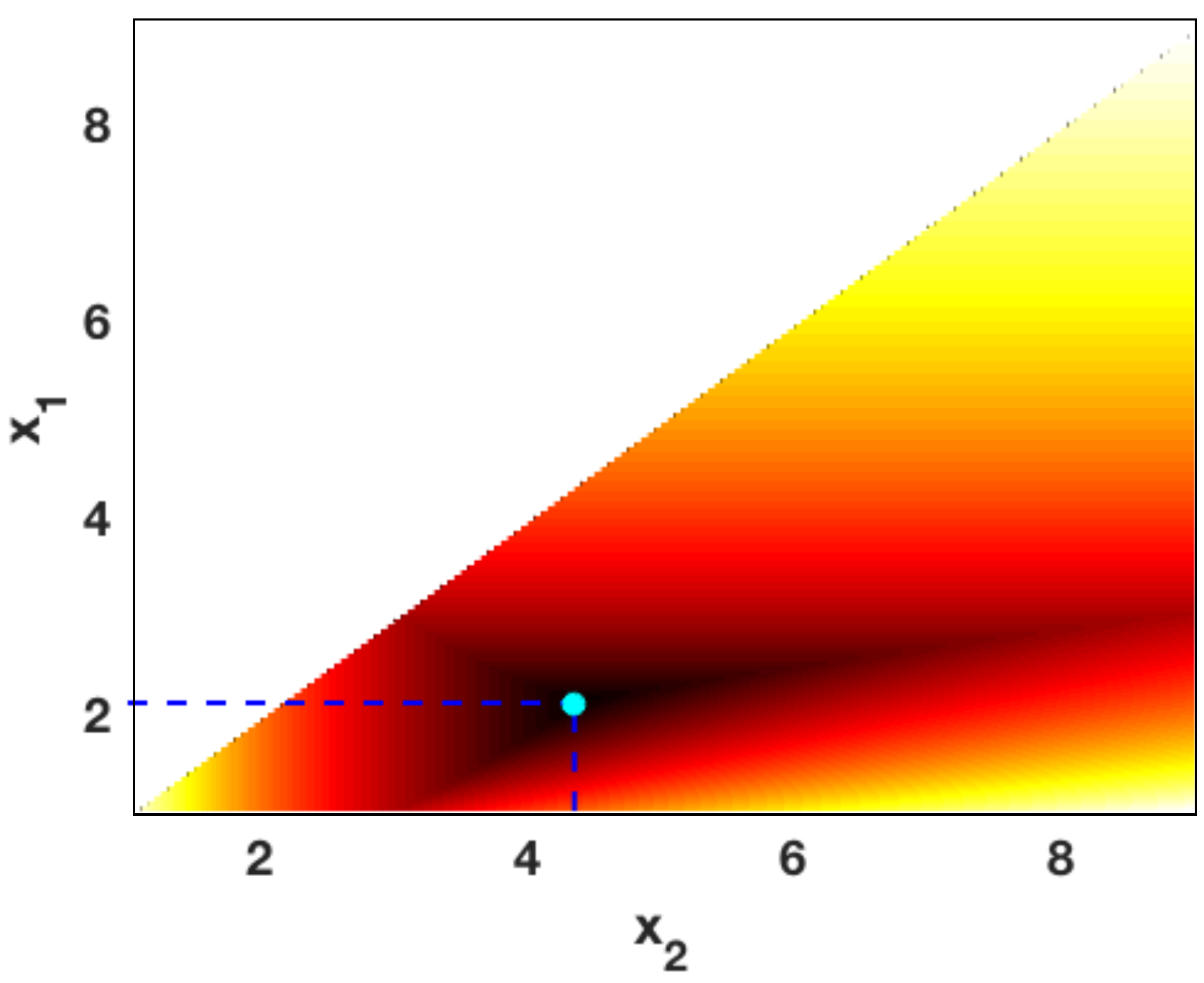}}
}
\caption{{\bf (a)} Optimal piecewise constant approximation $\phi_0(x)$ with $M=2$ nodes. {\bf (b)} The cost function $C_\infty(x_1,x_2)$ and its minimum at $(x_1^*,x_2^*)$. Note that $C_\infty(x_1,x_2)$ is defined within the simplex such that $x_1\leq x_2$, by definition; it can be also considered that $C_\infty(x_1,x_2)=C_\infty(x_2,x_1)$.}
\label{AppFig2}
\end{figure*}

\subsection*{Distributions of nodes }
\label{DistNodes}
We have seen that the auxiliary points $y_m$ are obtained using a uniform grid in the interval $[f(a),f(b)]$ (image of $[a,b]$). Therefore, $y_1,\ldots,y_M$ is a quasi-Monte Carlo sequence distributed uniformly in $[f(a),f(b)]$, i.e.,
\begin{equation}
y_m\sim \mathcal{U}([f(a),f(b)]), \qquad m=1,\ldots,M,
 \end{equation}
Following Eq. \eqref{ImpFormula}, we can find the distribution of the nodes $x_m$ since they are obtained by transforming the points $y_m$ through the function $f^{-1}(\cdot)$. Hence, following the expression of the transformation of a random variable,  we have 
\begin{eqnarray}
x_m\sim p_X(x) &=&p_Y(f(x)) \left|\frac{d f}{dx}\right|\\
   &\propto&\left|\frac{d f}{dx}\right|, \qquad m=1,\ldots,M,            
\end{eqnarray}
Therefore, the set of nodes $x_1,\ldots,x_M$ is a  quasi-Monte Carlo sequence with density $ p_X(x)\propto\left|\frac{d f}{dx}\right|$ and if $f$ is increasing, we can write $p_X(x)\propto\frac{d f}{dx}$. See Fig.~\ref{AppFig3} for an illustration of this.
For higher input dimension than 1 we have 
\begin{eqnarray}
{\bf x}_m\sim p_X({\bf x})\propto \left|\nabla f({\bf x})\right|.            
\end{eqnarray}

  \begin{figure*}[htbp]
\centering
\centerline{
\subfigure[]{\includegraphics[width=6.8cm]{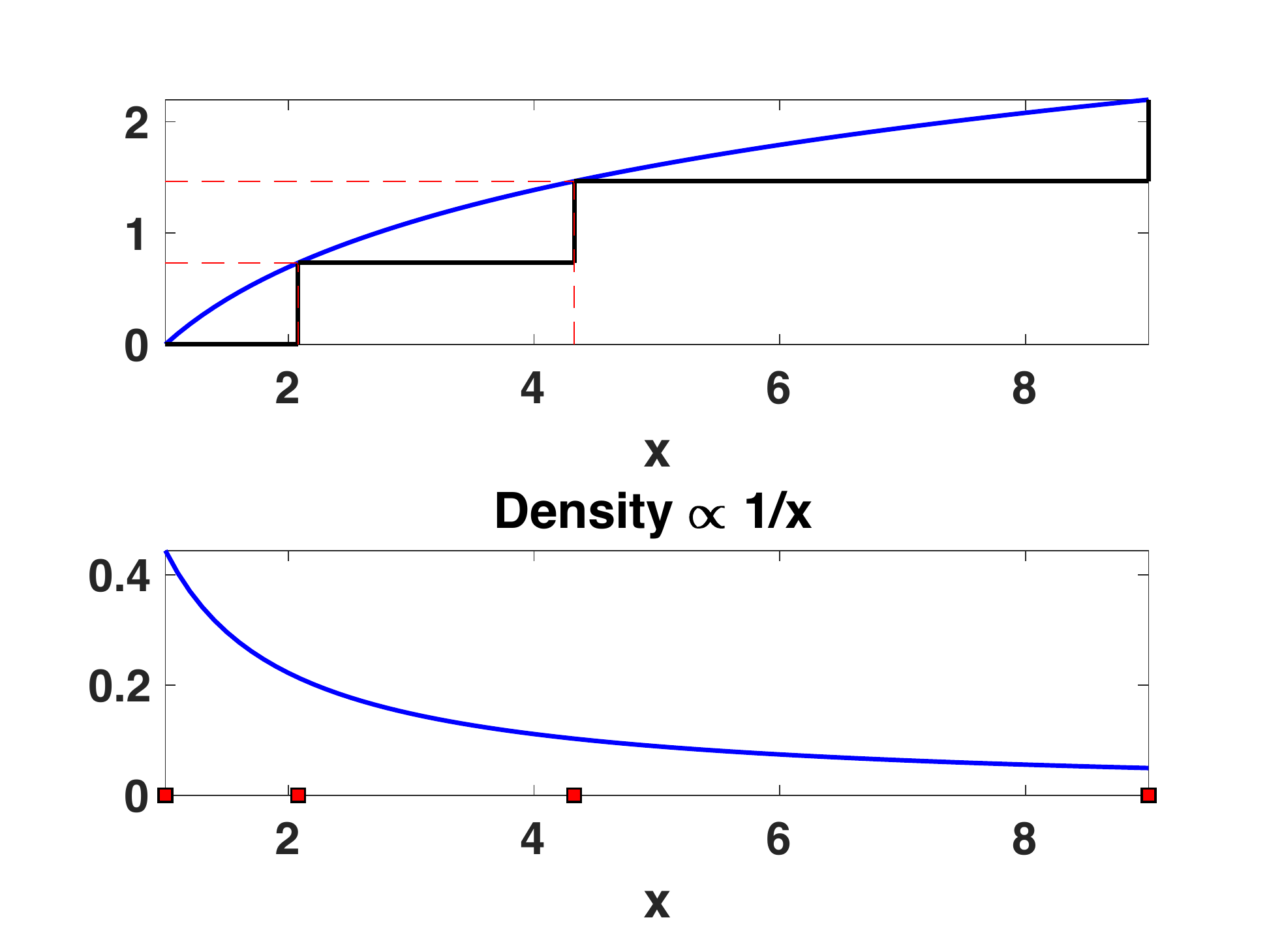}}
\subfigure[]{\includegraphics[width=7cm]{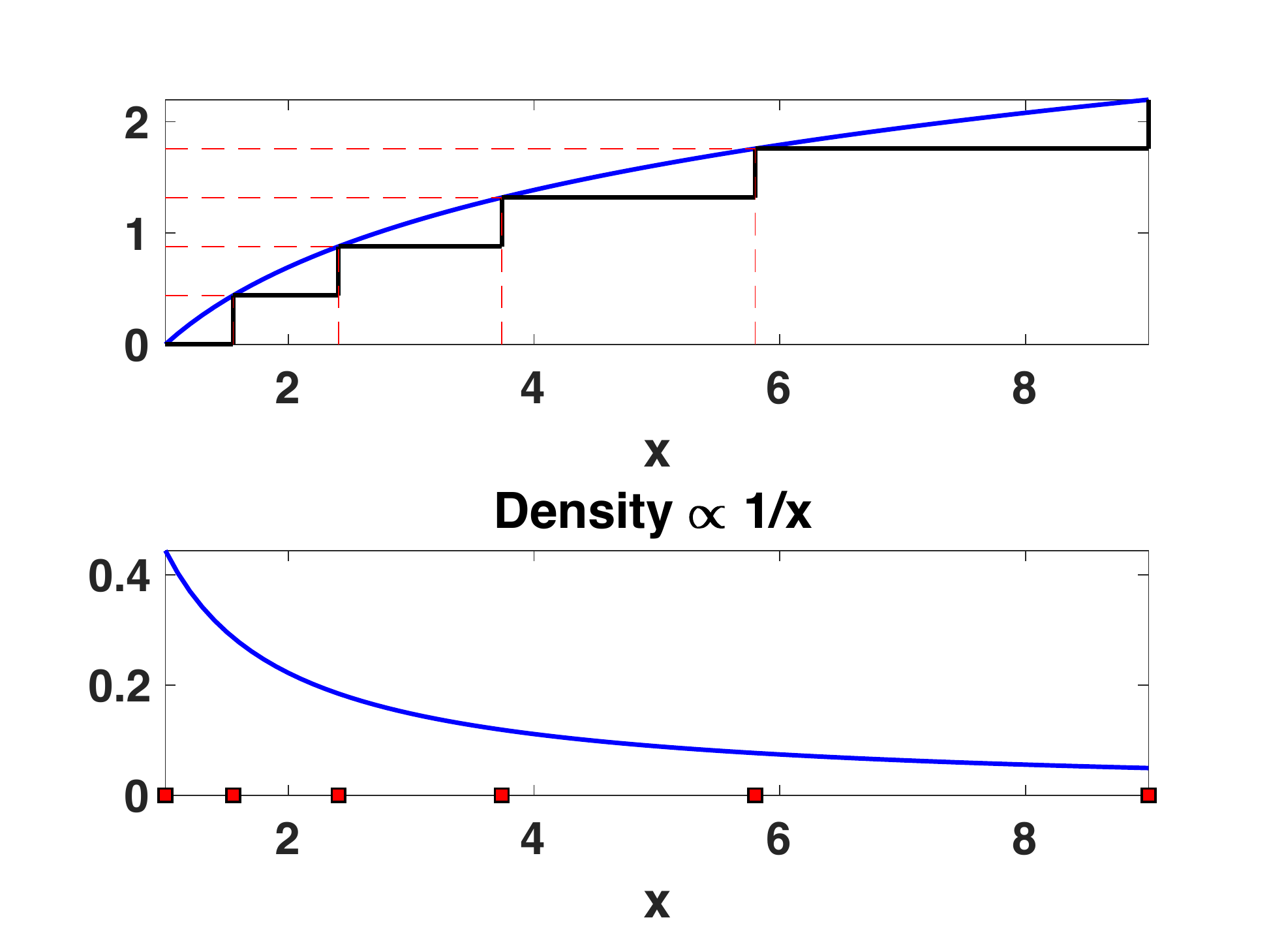}}
}
\centerline{
\subfigure[]{\includegraphics[width=6.8cm]{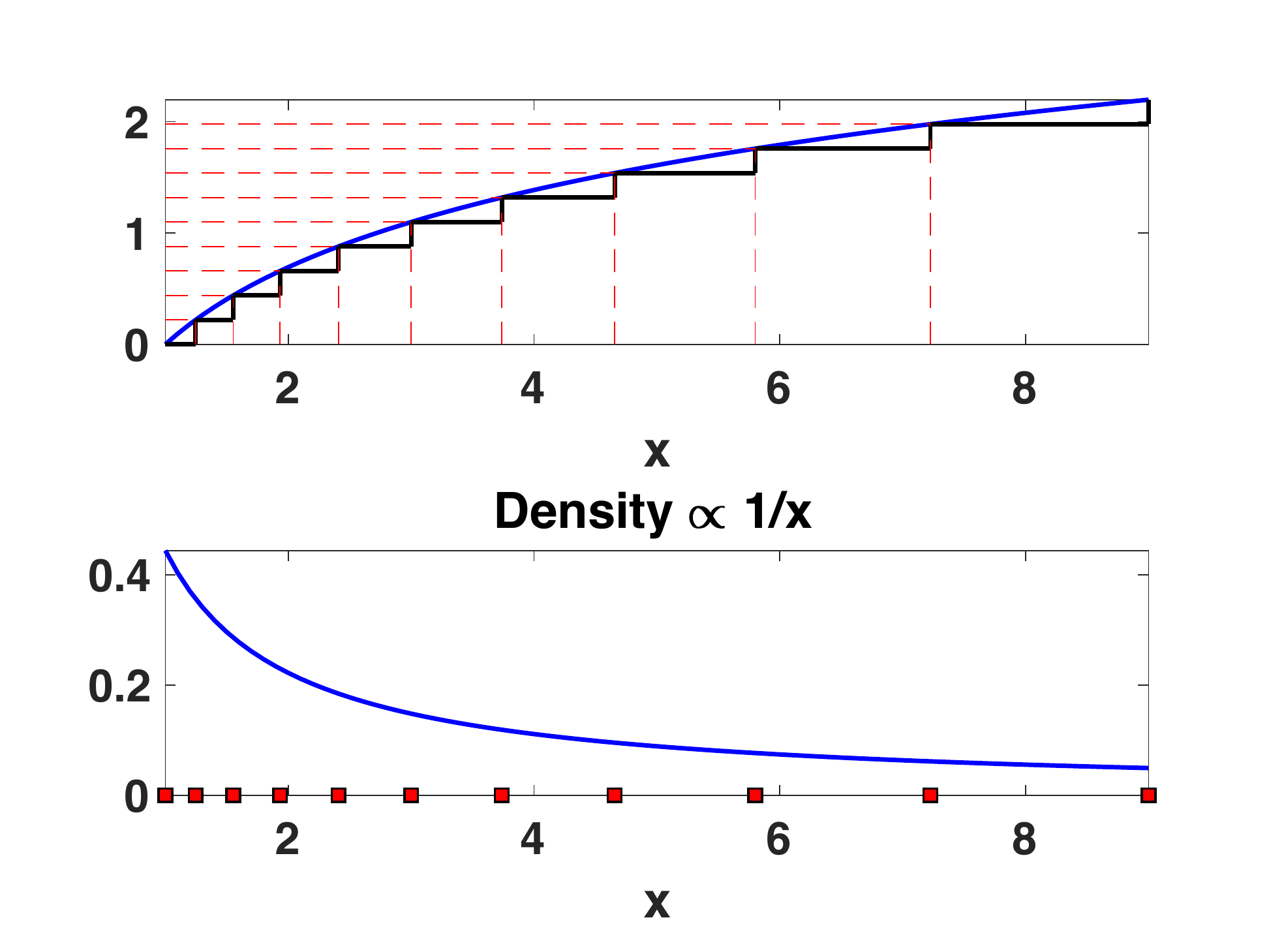}}
\subfigure[]{\includegraphics[width=7cm]{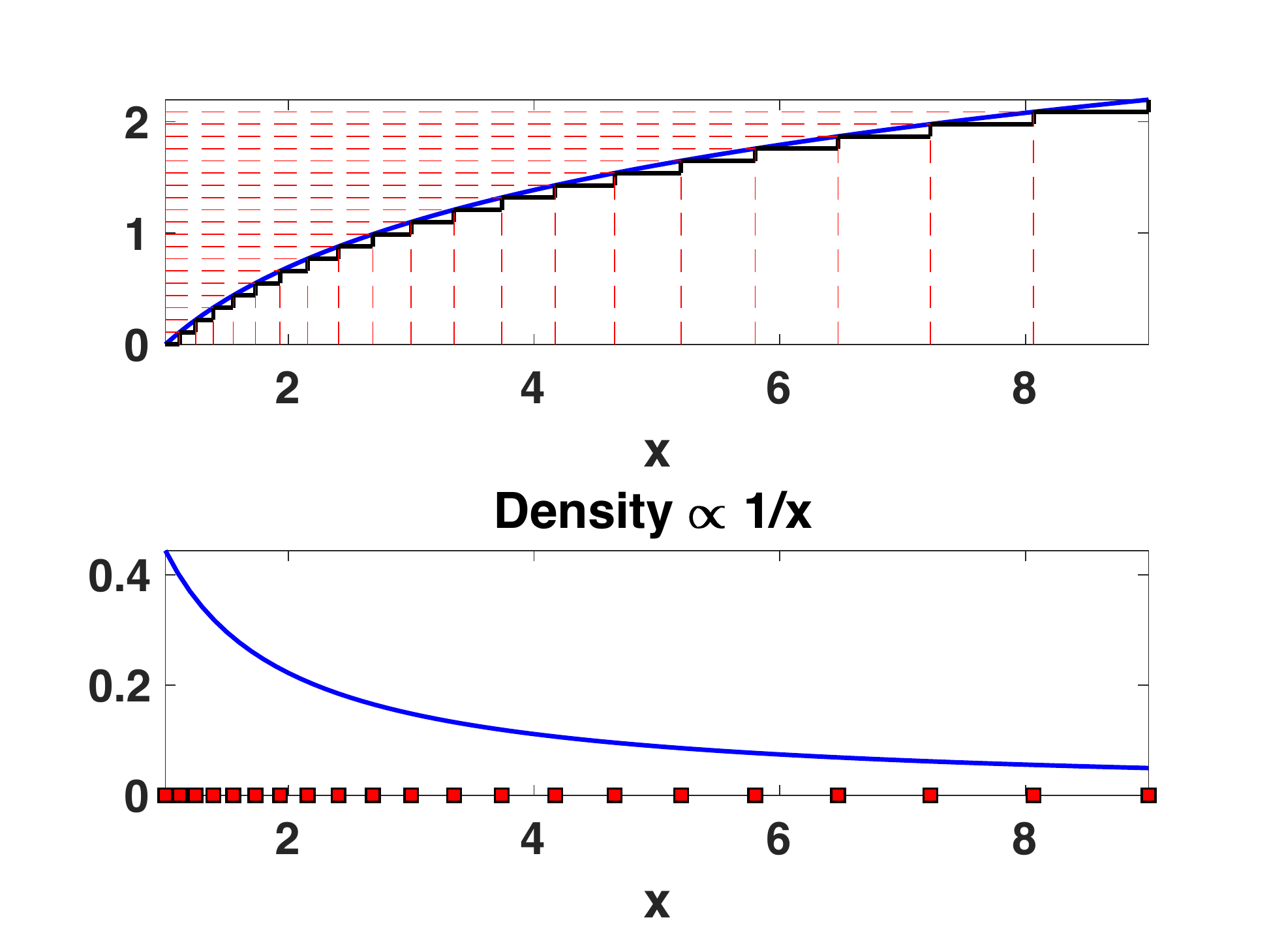}}
}
\caption{Illustration of function and optimal location of nodes (top) and density proportional to gradient (bottom). For (a)-(d) the number of nodes are 2, 4, 10 and 20 respectively.}
\label{AppFig3}
\end{figure*}

\end{document}